\documentclass[journal]{IEEEtran}

\usepackage{amsmath}
\usepackage{amsfonts}
\usepackage{amssymb}
\usepackage{amsthm}
\usepackage{pifont}
\usepackage[ruled,vlined]{algorithm2e}
\usepackage{algorithmic}
\usepackage{mathtools}
\usepackage{graphicx}
\usepackage{url}
\usepackage{verbatim}
\usepackage{cite}
\usepackage{bbding}
\usepackage{arydshln}

\graphicspath{{figures_diffusion/}}
\IEEEoverridecommandlockouts

\usepackage{array}
\usepackage{tabularx}
\usepackage{booktabs}
\usepackage{multirow}
\usepackage{textcomp}
\usepackage{setspace}
\usepackage{makecell}
\usepackage{threeparttable}
\usepackage{xcolor}

\definecolor{ground}{RGB}{0,0,255}
\definecolor{tree}{RGB}{0,255,0}
\definecolor{pole}{RGB}{255, 192, 203}
\definecolor{wire}{RGB}{255, 0, 0}

\newcommand{\topnum}[1]{%
	{\color[HTML]{FE0000}\textbf{#1}}%
}
\newcommand{\secondnum}[1]{%
	{\color[HTML]{DB3434}\textbf{#1}}%
} 
\newcommand{\thirdnum}[1]{%
	{\color[HTML]{F77B74}\textbf{#1}}%
}
\newcommand{\lastnum}[1]{%
	{\color[HTML]{3166FF}\textbf{#1}}%
}

\begin{document}

	\title{\LARGE \bf Sem-RaDiff: Diffusion-Based 3D Radar Semantic Perception \\ in Cluttered Agricultural Environments}
	\author{Ruibin Zhang and Fei Gao

		\thanks{
		All authors are with the Institute of Cyber-Systems and Control, College of Control Science and Engineering, Zhejiang University, Hangzhou 310027, China, and also with the Huzhou Institute, Zhejiang University, Huzhou 313000, China.}
		\thanks{Corresponding author: Fei Gao.}
		\thanks{Email:  \tt \small  \{ruibin\_zhang, fgaoaa\}@zju.edu.cn}
		\thanks{This work was supported by the National Natural Science Foundation of China under grant no. 62322314 and the Fundamental Research Funds for the Central Universities.}
	}

	\maketitle
	\thispagestyle{empty}
	\pagestyle{empty}

	\begin{abstract}
		\label{sec:abstract}\textbf{
         	Accurate and robust environmental perception is crucial for robot autonomous navigation. While current methods typically adopt optical sensors (e.g., camera, LiDAR) as primary sensing modalities, their susceptibility to visual occlusion often leads to degraded performance or complete system failure. In this paper, we focus on agricultural scenarios where robots are exposed to the risk of onboard sensor contamination. Leveraging radar's strong penetration capability, we introduce a radar-based 3D environmental perception framework as a viable alternative. It comprises three core modules designed for dense and accurate semantic perception:
         	1) Parallel frame accumulation to enhance signal-to-noise ratio of radar raw data.
         	2) A diffusion model-based hierarchical learning framework that first filters radar sidelobe artifacts then generates fine-grained 3D semantic point clouds.
         	3) A specifically designed sparse 3D network optimized for processing large-scale radar raw data. 
         	We conducted extensive benchmark comparisons and experimental evaluations on a self-built dataset collected in real-world agricultural field scenes. Results demonstrate that our method achieves superior structural and semantic prediction performance compared to existing methods, while simultaneously reducing computational and memory costs by 51.3\% and 27.5\%, respectively. Furthermore, our approach achieves complete reconstruction and accurate classification of thin structures such as poles and wires\textemdash which existing methods struggle to perceive\textemdash highlighting its potential for dense and accurate 3D radar perception.
		}
		
	\end{abstract}

	\IEEEpeerreviewmaketitle

    \section{Introduction}
    \label{sec:introduction}	
	The development of agricultural machinery has been intertwined with human history. Since the Industrial Revolution, inventions such as tractors have significantly increased production efficiency and contributed to the growth of the world population. 
	Nowadays, we are able to witness agricultural robots busy in the fields with full autonomy, which further increase productivity and save manpower. For instance, some agricultural drones' automatic spraying efficiency are approximately 70 times higher than manual spraying\cite{wakchaure2023application}. Such achievements are mainly attributed to the advancement of autonomous navigation technology. Among them, environmental perception is crucial, as it provides environmental semantic and structural information to assist robots in simultaneous localization and mapping (SLAM), as well as subsequent decision-making and planning. Existing onboard sensing modalities primarily include camera and LiDAR sensors, with the former offering dense semantic information at a low cost, and the latter providing accurate range measurements and a larger sensing range.  However, they are all susceptible to performance degradation due to occlusion by tiny particles or dirt, even leading to complete failure of autonomous missions. In agricultural scenarios, onboard sensors are easily soiled, which poses a great threat to the safety of agricultural robots.
	

	\begin{figure}
		\centering
		\includegraphics[width=1\linewidth]{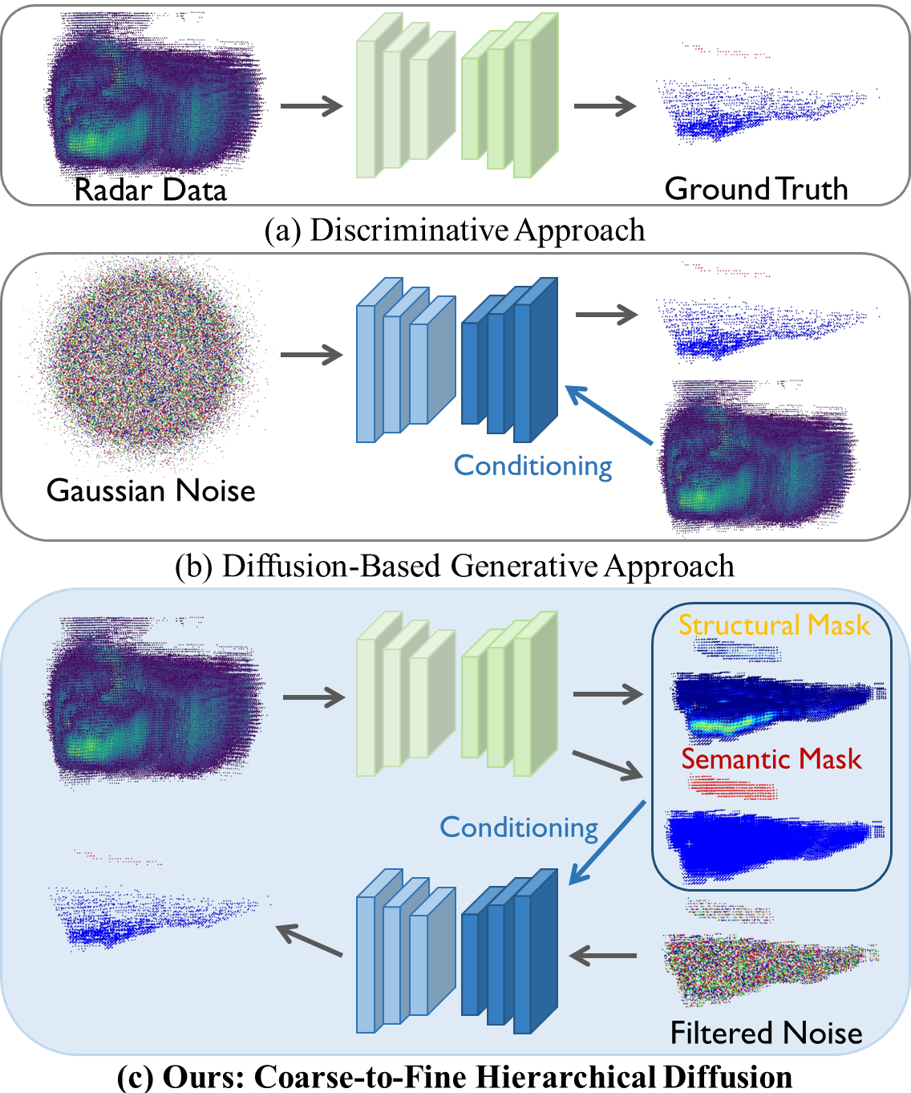}
		\caption{
			\textbf{Comparison of different approaches to cross-modal radar perception}. (a) Discriminative approaches attempt to make radar resemble high-performance sensors through  supervised learning. (b) Generative approaches take radar data as conditional input to achieve perception enhancing utilizing generative learning strategies such as diffusion models. (c) We integrate the above approaches into a coarse-to-fine paradigm that first reduce the modality gap by filtering massive and noisy radar data into interpretable coarse-grained masks and then predict fine-grained 3D semantic point clouds.
		}
		\label{pic:topgraph}
		\vspace{-0.5cm}
	\end{figure}

	To address this issue, both the academia and industry communities are exploring radar, especially millimeter-wave (mmWave) radar, as a complementary sensing modality, which has already been deployed in a number of commercial agricultural robots\cite{xag2021robots, eavision2021robots}. Compared to camera and LiDAR sensors, it operates at millimeter-wave band and therefore possesses much stronger penetration capability, making it robust in various weather conditions, and is expected to replace these optical sensors in agricultural scenarios. However, it suffers from inherent defects such as low signal-to-noise ratio (SNR) and low angular resolution. Traditional radar target detectors, such as CFAR\cite{richards2005fundamentals} and MUSIC\cite{schmidt1986multiple}, have difficulty in distinguishing between noise and valid targets in raw mmWave radar data, as well as depicting the structure of valid targets. Applying these algorithms can only yield sparse point clouds, accompanied by the risks of miss detections and false alarms. This severely constrains the performance of downstream tasks such as ego-motion estimation\cite{cen2018precise, kramer2020radar, almalioglu2020milli}, object detection\cite{gao2020ramp, wang2021rodnet, cheng2021robust}, and semantic segmentation\cite{schumann2018semantic, kaul2020rss, zeller2023radar, dalbah2024transradar}. Therefore, mmWave radar is now usually merely used as an auxiliary sensing modality.
	To enhance mmWave radar perception performance, recent studies attempt to extract denser and more accurate point clouds from mmWave radar data via learning-based approach. As stated above, mmWave radar suffers from various drawbacks, which can be divided into the following three aspects that pose great challenges for these studies.
	
	
	\begin{itemize}
		\item [1)] \textbf{Low SNR and susceptibility to interference}: In cluttered environments, the mmWave radar echo energy of valid targets is greatly attenuated due to multiple scattering. The power intensity of some targets with low radar cross-section (RCS) is even lower than the noise intensity, making them difficult to distinguish. Additionally, mmWave radar is susceptible to interference from multipath effects and ground reflections, further polluting the radar return signal. Agricultural environments are filled with dense targets such as crops and trees, posing significant challenges for radar perception.
		
		\item [2)] \textbf{Low angular resolution and sidelobe contamination}: Radar's angular resolution depends on its antenna aperture. For mmWave radar, a meter-level antenna aperture is required to achieve the same angular resolution as LiDAR\cite{rao2017introduction}, which is impractical for deployment on agricultural robots. With a much lower angular resolution, mmWave radar struggles to distinguish nearby targets and accurately depict their structures. In addition, sidelobe artifacts in the echo signals from prominent targets may drown out other signals, resulting in missed detections. In agricultural environments, small targets such as power lines and poles are prone to being submerged by ground echoes, causing safety hazards.

		\item [3)] \textbf{Massive mmWave radar data scale}: The raw data of mmWave radar is presented as spatial spectrum in tensor form, which implies huge data scale. 4D radar (with the additional dimension of Doppler velocity measurements) data can occupy up to 260MB in a single frame\cite{paek2022k}.
		The computational overhead when processing data of such a scale is prohibitive for onboard computers. Furthermore, the low SNR and low angular resolution make it difficult to filter out noise to reduce the data scale.
	\end{itemize}

    \begin{figure*}[t]
	\centering
	\includegraphics[width=1.0\linewidth]{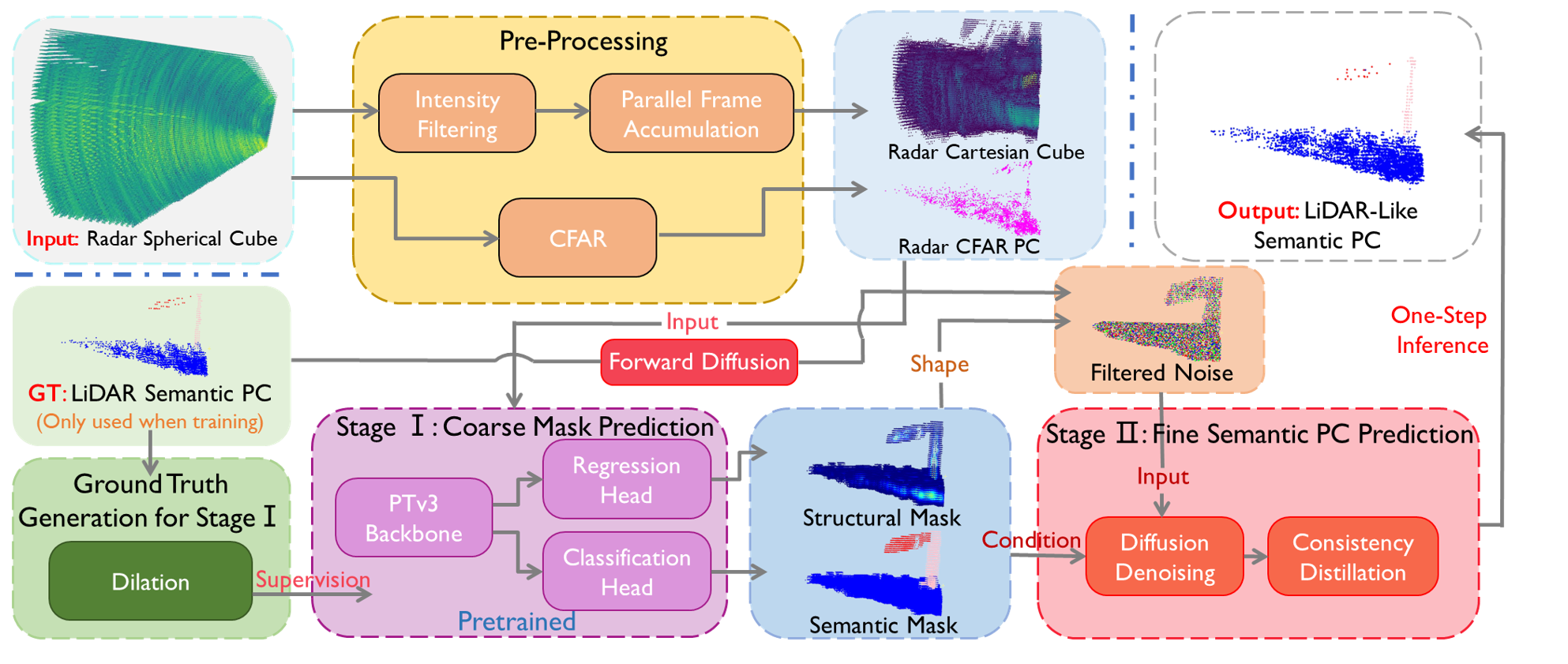}
	\caption{
		Overall framework of our Sem-RaDiff. It consists of three modules: (1) Radar data pre-processing that transforms multi-frame radar spherical cubes into accumulated radar Cartesian cube and radar CFAR PC. (2) Coarse mask prediction that predicts structural and semantic masks under the supervision of dilated LiDAR semantic PC. (3) Fine semantic PC prediction that generates LiDAR-like semantic PC via diffusion model and one-step consistency sampling, conditioned on the structural and semantic masks. PC is short for point clouds.
	}
	\label{pic:architecture}
	\vspace{-0.4cm}
	\end{figure*}

	As far as we know, existing methods fail to solve all of the aforementioned problems. Several approaches simply formulate radar perception enhancement as discriminative learning\cite{cheng2021new, cheng2022novel, ronneberger2015u, fan2024enhancing, geng2024dream, xu2022learned, kim2024pillargen, prabhakara2023high, hunt2024radcloud, han2024denserradar, roldan2024deep}, directly learning the conditional distribution between radar data and the ground truth (primarily LiDAR point clouds). Unlike common tasks such as image semantic segmentation, there is no obvious mapping between inherently noisy radar data and accurate ground truth. Consequently, such methods inevitably suffer from underfitting, resulting in network predictions that remain substantially inferior to the ground truth. Alternative methods adopt generative learning paradigms (e.g., diffusion models) to instead learn the joint distribution of the conditioning input (radar data) and the ground truth\cite{lu2020see, guan2020through, sun20213drimr, sun2021deeppoint, zhang2024towards, wu2024diffradar, luan2024diffusion, wu2025diffusion, zheng2025r2ldm, yang2025unsupervised, wang2025sddiff}. While these methods can synthesize high-quality, ground-truth-like point clouds through their generative capability, they face significantly greater training difficulties, such as mode collapse or overfitting. The high-dimensional and large-scale nature of radar data further exacerbates these challenges, while simultaneously imposing prohibitive computational demands that hinder onboard deployment. To circumvent these difficulties, current solutions typically adopt compromised strategies: either projecting data into Bird's-Eye-View (BEV) representations or solely utilizing radar point clouds generated by traditional detectors (RPC) as conditioning inputs - both approaches incur substantial information loss that critically degrades performance. In summary, existing methods have inherent flaws in the design of learning paradigms, data modalities or representation spaces, rendering them particularly inadequate for complex real-world scenarios such as agricultural environments where precision and robustness are paramount.
	
	
	In this paper, we present a coarse-to-fine diffusion model-based approach called \textbf{Sem-RaDiff} (as shown in Fig.\ref{pic:topgraph} and Fig.\ref{pic:architecture}) that addresses the issues above. The proposed method can generate high-quality 3D radar point clouds along with semantic labels to facilitate the subsequent tasks of agricultural robots. To address the noise characteristics of mmWave radar, we propose a parallel frame accumulation algorithm that performs real-time non-coherent accumulation of current-frame radar raw data with past frames to enhance SNR. In view of the low angular resolution and sidelobe contamination issues, we propose a two-stage learning framework. In the first stage, the network learns to filter out sidelobe artifacts from radar data using dilated LiDAR point clouds as supervision, producing coarse-grained intermediate representations. These preliminary results then serve as conditional inputs for the second stage, where a diffusion model refines them into fine-grained 3D LiDAR-like semantic point clouds with enhanced geometric and semantic fidelity. 
	Moreover, to handle large-scale millimeter-wave radar data while overcoming the limitations of prior methods that either reduce dimensionality or discard raw radar measurements, we introduce a sparse 3D representation which preserves complete 3D structural information while significantly reducing computational overhead and memory consumption. We accordingly design a sparse 3D network that performs coarse mask prediction and fine point cloud generation for both learn stages.
	For practical onboard deployment, we also integrate inference acceleration techniques for diffusion models to enable one-step generation, while maintaining generation quality.
	To verify and evaluate the proposed method, we take an agricultural drone as the carrier, build a dataset in agricultural fields, and carry out extensive benchmark comparisons and ablation studies, confirming that the proposed method demonstrates state-of-the-art performance.
	Detailed contributions are as follows.

	\begin{itemize}
		\item [1)] 
		Introducing Sem-RaDiff, the first systematic solution for simultaneous mmWave radar perception enhancement and semantic segmentation in cluttered agricultural environments. Featured parallel frame accumulation and hierarchical diffusion-based learning paradigm, Sem-Radiff enables predicting LiDAR-like radar semantic point clouds in a coarse-to-fine manner.
		
		\item [2)]
		Proposing a sparse representation along with a dedicated sparse 3D network for higher computational efficiency without performance degradation. Considering onboard deployment requirements, we further incorporate one-step inference technique for diffusion models, eliminating reliance on iterative sampling.
		
		\item [3)]
		We conduct comprehensive benchmark comparisons and ablation studies on a self-built dataset recorded with a commercial agricultural drone. The results, along with real-world validation, demonstrate the state-of-the-art performance of the proposed solution. 

	\end{itemize}



    \section{Related Work} 
    \label{sec:related_works}	
    \subsection{Discriminative Radar Point Cloud Generation} 
    \label{sec:related_works_discriminative}	
    Discriminative learning-based methods have been extensively studied for radar point cloud generation\cite{cheng2021new, cheng2022novel, ronneberger2015u, fan2024enhancing, geng2024dream, xu2022learned, kim2024pillargen, prabhakara2023high, hunt2024radcloud, han2024denserradar, roldan2024deep}. Cheng et al.\cite{cheng2021new, cheng2022novel} first propose RPDNet, a learning framework based on U-net\cite{ronneberger2015u} architecture that filters falsely detected points in the range-Doppler matrix (RDM) generated by range FFT and Doppler FFT under the supervision of LiDAR point clouds, and then employ conventional Direction of Arrival (DOA) estimation algorithm to generate 3D radar point clouds. Fan et al.\cite{fan2024enhancing} further introduces visual-inertial supervision instead of LiDAR supervision based on RPDNet framework. These methods still rely on conventional DOA estimation with limited angular resolution. DREAM-PCD\cite{geng2024dream} integrates both incoherent and coherent accumulation techniques to enhance radar point cloud density and angular resolution respectively, then perform noise-filtering via neural network. Coherent accumulation requires sub-millimeter level pose estimation to achieve prominent resolution improvement, which is infeasible in robotic applications. Consequently, the actual performance gain from coherent accumulation is constrained, yielding point clouds that remain noise-corrupted and fail to accurately recover object structures. For computational efficiency, some methods encode both RPC and ground truth into depth images\cite{xu2022learned} or pillar representations\cite{kim2024pillargen}. The above approaches remain constrained by conventional radar signal processing algorithms, and the resulting point clouds are still inferior to ground truth. 
    
    Alternative approaches employ end-to-end frameworks that utilize neural networks to directly extract point clouds from raw radar data. RadarHD\cite{prabhakara2023high} and RadCloud\cite{hunt2024radcloud} take the range-azimuth heatmaps (RAHs) as input to generate 2D BEV point clouds under LiDAR supervision. Han et al.\cite{han2024denserradar} and Roldan et al.\cite{roldan2024deep}, on the other hand, take the 4D radar cubes (incorporating the additional Doppler dimension) as input, generating 3D point clouds via neural networks based on computationally expensive 3D convolutional neural network (CNN). Discriminative models learn a smooth mapping between inputs and outputs, but due to the substantial modality gap between the bulky and low-resolution raw radar data and the fine-grained ground truth, such methods suffer from severe underfitting. Whether in training or test sets, sidelobe artifacts within the raw data cannot be completely suppressed, often resulting in the merging of multiple targets in the generated point clouds, making them difficult to differentiate. 
    
    
    \subsection{Generative Radar Point Cloud Generation} 
    \label{sec:related_works_generative}	
    Various approaches have been proposed to achieve high-quality radar point cloud generation via generative models \cite{lu2020see, guan2020through, sun20213drimr, sun2021deeppoint, zhang2024towards, wu2024diffradar, luan2024diffusion, wu2025diffusion, zheng2025r2ldm, yang2025unsupervised, wang2025sddiff}. Early attempts typically adopt conditional generative adversarial networks (cGANs)\cite{mirza2014conditional} that use adversarial training to learn the joint distribution of paired radar data and ground truth. Millimap\cite{lu2020see} learns BEV point clouds for indoor environments from RPCs utilizing cGAN and geometric priors of indoor spaces under LiDAR supervision. Hawkeye\cite{guan2020through}, 3DRIMR\cite{sun20213drimr}, and DeepPoint\cite{sun2021deeppoint} achieve 3D reconstruction for specified objects from 3D radar cubes. However, the adversarial training characteristics of cGANs can lead to training instability and mode collapse issues, hence limiting the practicality of cGAN-based radar point cloud generation to predefined scenarios with limited diversity.
    
    Recent advances in diffusion models\cite{ho2020denoising, song2020score} have unlocked new possibilities for radar point cloud generation. Characterized by progressive generation and simple training objectives, diffusion models demonstrate superior training stability and sample diversity compared to GANs\cite{goodfellow2014generative} and Variational Autoencoders (VAEs)\cite{kingma2013auto}. In our previous work, Radar-Diffusion\cite{zhang2024towards}, we first introduce diffusion models to radar point cloud generation, demonstrating potential for generating LiDAR-like BEV point clouds from RAHs. Other related often employ RPCs as the conditional input, attempting to generate LiDAR-like point clouds via diffusion models across diverse data representations—such as BEV images\cite{wu2024diffradar}, elevation maps\cite{luan2024diffusion}, depth images\cite{wu2025diffusion}, and 3D voxel grids\cite{zheng2025r2ldm}. Yang et al.\cite{yang2025unsupervised} proposed an unsupervised scheme that leverages a diffusion model guided by arbitrary LiDAR domain knowledge rather than paired RPCs and LiDAR point clouds. The aforementioned methods either use RAH or RPC, rather than 3D/4D radar cubes, as conditional inputs, leading to a significant loss of information. SDDiff\cite{wang2025sddiff} proposes the use of 4D radar cubes as conditional inputs, achieving both point cloud generation and ego-motion estimation. However, it neglects the issues of low SNR and low angular resolution in 4D radar cubes, resulting in miss detections of weak-echo targets in the scene. Additionally, it relies on expensive 3D CNNs, making it challenging for onboard deployment.
    
    To address these drawbacks, we propose parallel frame accumulation (Section \ref{sec:preprocessing}) and neural radar cube filtering (Section \ref{sec:stage1}) to effectively handle the SNR and low angular resolution issues in raw radar cubes. Additionally, we introduce a sparse 3D network and incorporate one-step inference technique to facilitate onboard deployment, with the aim of achieving real-time, high-quality 3D radar point cloud generation in cluttered agricultural scenarios.

    
    \subsection{Radar-only Semantic Prediction} 
    \label{sec:related_works_semantic}	
    Several works explore semantic prediction using only radar sensor\cite{schumann2018semantic, zeller2023radar, zeller2024semrafiner, ma2024licrocc, ding2024radarocc}. Schumann et al.\cite{schumann2018semantic} propose a semantic segmentation algorithm for RPCs based on PointNet++\cite{qi2017pointnet}, specifically designed for automotive scenarios. Zeller et al.\cite{zeller2023radar, zeller2024semrafiner} present a systematic approach for moving instance segmentation on RPCs utilizing temporal information via a sequential attentive feature encoding
    module. However, the semantic information in sparse and noisy RPCs is extremely limited. For example, both pedestrian and bicycle objects may manifest as merely a few scattering points, rendering them hardly indistinguishable. This limitation imposes fundamental constraints on the performance of these methods. Ding et al.\cite{ding2024radarocc} and Sun et al.\cite{sun2025automatic} instead propose using 4D radar cubes for 3D semantic scene completion via cross-modal supervision. Similar to radar point cloud generation methods based on discriminative models (as discussed in Section \ref{sec:related_works_discriminative}), the inherent flaws of 4D radar cubes result in low resolution and the presence of artifacts in the generated semantic voxels. Crucially, suboptimal radar data quality intrinsically constrains the performance regardless of input representation (RPCs or 4D radar cubes). We address this problem by simultaneously achieving radar perception enhancement and semantic prediction. In Stage {\uppercase\expandafter{\romannumeral1} of Sem-RaDiff, the network outputs coarse-grained  structural and semantic masks, which jointly serve as conditional inputs for Stage {\uppercase\expandafter{\romannumeral2} to generate fine-grained 3D semantic point clouds.
    
    \subsection{Robot Perception in Agricultural Environments}
	\label{sec:background}
	In recent years, agricultural robots have been capable of undertaking tasks such as seeding, harvesting, irrigation and pesticide spraying \cite{wakchaure2023application}. Many commercial agricultural robot products with autonomous navigation capabilities are now available\cite{xag2021robots, eavision2021robots}. Agricultural robots often operate around splashing pesticides, water droplets, crop debris, etc., and thus the onboard sensors are frequently soiled, as shown in Fig.\ref{pic:Background}. As seen in the figure, when covered with mud, the field of view (FOV) of camera and LiDAR sensors is severely blocked, causing the drone to be "blind". In contrast, mmWave radar is almost unaffected, but due to its inherent sensing limitations, a large number of miss detections and false alarms occur in both conditions. The aim of this work is to improve the quality of mmWave radar point clouds via data-driven approach, as well as to endow them with semantic features, in agricultural field scenarios.

	\begin{figure}
		\centering
		\includegraphics[width=1\linewidth]{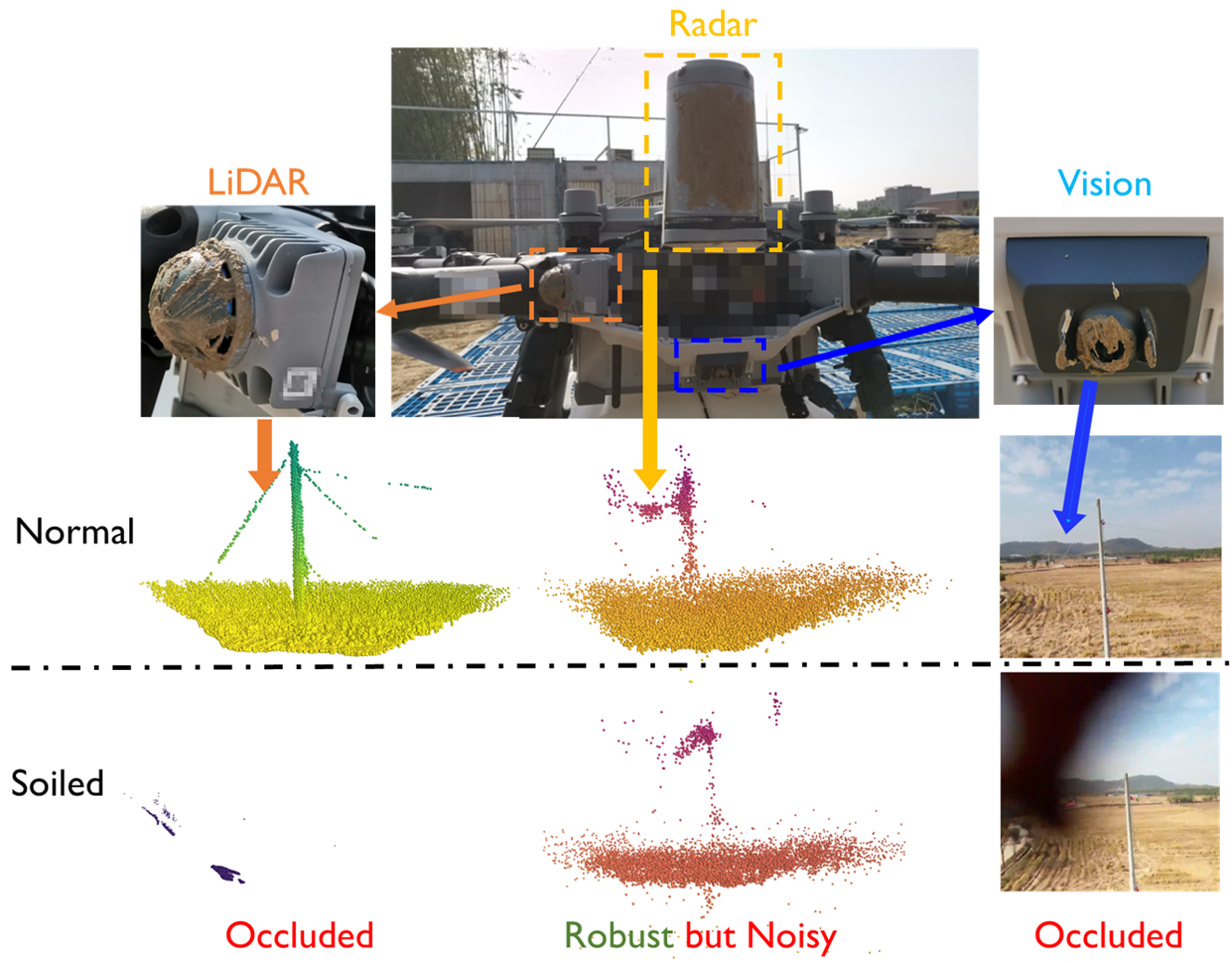}
		\caption{
			Comparison of different sensing modalities under normal and soiled conditions in agricultural settings. 
		}
		\label{pic:Background}
			\vspace{-0.5cm}
	\end{figure}

    \section{Preliminaries}
    \label{sec:Preliminaries}

 	\begin{figure}
 	\centering
 	\includegraphics[width=1.0\linewidth]{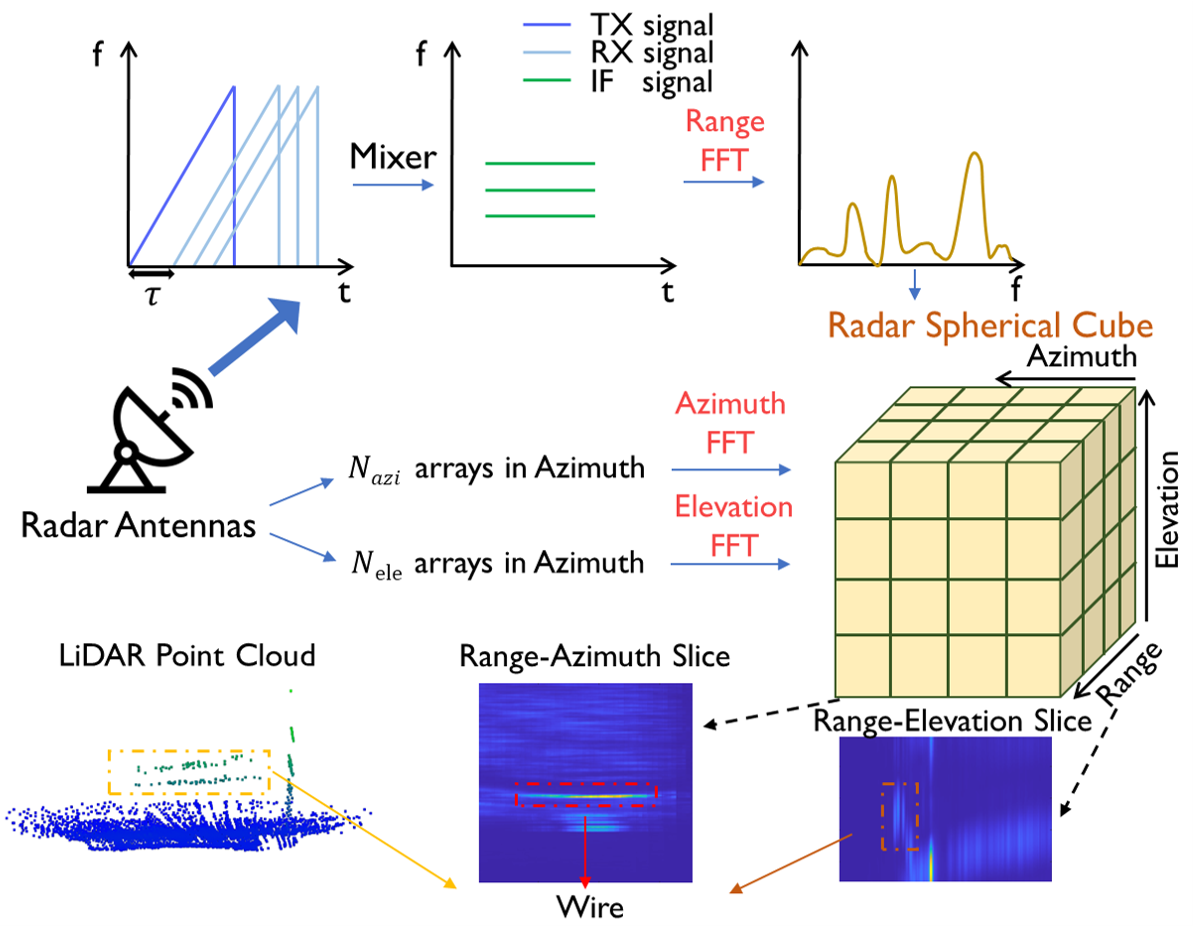}
 	\caption{
 		Illustration of the radar signal processing pipeline, as well as the visual comparison between radar spherical cube and LiDAR point cloud.
 	}
 	\label{pic:radar_principle}
 	\vspace{-0.4cm}
 	\end{figure}

    \subsection{Radar Signal Processing}
 	\label{sec:radar_signal_processing}
 	Mmwave radar typically operates based on frequency-modulated continuous wave (FMCW) technology. This involves the radar continuously transmitting electromagnetic chirps whose frequency vary over time. The range, velocity, and Bearing of objects are then determined by analyzing the receive (RX) signals from those objects. Linear modulation is predominantly adopted where the frequency of the transmit (TX) signal increases linearly with time\cite{gao2021mimo}:
 	\begin{equation}
 	s_T(t)=A_T\exp \left[j 2 \pi\left(f_0 t+\frac{1}{2} \frac{B}{T_c} t^2\right)\right], t \in[0, T_c], 
	\end{equation}
 	where $A_T$ denotes the amplitude, $f_0$ denotes carrier frequency, $B$ denotes bandwidth, $T_c$ denotes chirp duration. The RX signal is a time-delayed version of TX signal, i.e., $S_R(t) = S_T(t-\tau), \tau = \frac{2r}{c}$, where $r$ is the range of object from radar sensor and $c$ is the speed of light. A mixer then calculates the difference of $s_T$ and $S_R$ with the resulting signal called intermediate frequency (IF) signal. Linear modulation waves hold the property that the frequency of IF signal remains constant and is equal to $\frac{B}{T_c}\tau$. The IF signal is then processed by analog-to-digital converter (ADC) to obtain discrete samples.
 	After applying fast Fourier transform (FFT) to the IF signal, each peak (assumed to be located at frequency $f_i$) of the IF signal indicates a detected object, with range given by $r_i = f_i\frac{c T_c}{2B}$ and resolution given by $\frac{c}{2B}$. For commercial mmWave radar systems, the typical bandwidth of 1-4 GHz yields a corresponding range resolution of 3.75 cm to 15 cm.
 	
 	Velocity measurement is based on the analysis of Doppler effect occurs in consecutive chirps. However, in scanning radars, including the one employed in this work, sensor rotation prevents reliable estimation of Doppler velocity. We will demonstrate that high-quality 3D environmental perception can still be achieved without velocity information.
 	
 	Bearing (azimuth and elevation) estimation relies on analyzing the phase shifts across spatially separated RX antennas. Each object induces distinct phase differences across the RX antennas. Utilizing multiple-input and multiple-out (MIMO) technology,  $N_{T}$ TX antennas and $N_{R}$ RX antennas spaced at intervals 
 	$d$ can form a $N_a = N_{T} \times N_{R}$ virtual array. Performing FFT on the sequence of phasors yields spectral peaks (assumed to be located at phase $\phi_i$) corresponding to detected objects, with bearing angle given by $\theta_i = sin^{-1}(\frac{\lambda \phi_i}{2\pi d})$ and angular resolution given by $\frac{\lambda}{N_a d}$.
 	
 	
 	In this work, we perform FFTs on the time-domain radar signal across range, azimuth, and elevation dimension, yielding a 3D data cube representing the spatial spectrum in spherical coordinates (termed the radar spherical cube), as shown in Fig.\ref{pic:radar_principle}. Each bin within this cube stores echo intensity, where peaks indicate potential targets. These detections, however, are degraded by factors like low SNR and limited angular resolution. In this work, we aims at performing 3D environmental perception and semantic prediction from radar spherical cubes via the proposed Sem-RaDiff. 
 	

	\subsection{Diffusion Models}
	\label{sec:diffusion_models}
 	Diffusion models are a class of probabilistic generative model that progressively corrupts data samples to noise by Gaussian perturbation kernels (called forward diffusion process), and learn to sequentially recover the original samples (called reverse diffusion process), thereby learning the underlying data distribution $p_{\text {data}}(\boldsymbol{x})$. From the perspective of continuous-time diffusion models\cite{song2020score, karras2022elucidating}, the forward diffusion process is achieved through a forward stochastic differential equation (SDE) defined as follows:
 	\begin{equation}
 	\mathrm{d} \boldsymbol{x}=\boldsymbol{f}(\boldsymbol{x}, \sigma) \mathrm{d} \sigma+g(\sigma) \mathrm{d} \omega_\sigma, 
 	\end{equation}
	where $\sigma \in[0, \sigma_{max}]$ is a continuous variable which represents the noise level, $ \omega_\sigma$ is a standard Wiener process, $\boldsymbol{f}(\cdot, \sigma)$ is a vector-valued function called the drift coefficient, and $g(\cdot)$ is a scalar function called the diffusion coefficient. For example, in the settings of EDM\cite{karras2022elucidating}, $\boldsymbol{f}(\boldsymbol{x}, \sigma) = \boldsymbol{0}$ and $g(\sigma) = \sqrt{2\sigma}$, then we have $p(\boldsymbol{x}_{\sigma}\mid\boldsymbol{x}) \sim\mathcal{N}(\boldsymbol{x}, \sigma^2 \mathbf{I})$. 
	
	The reserve diffusion process generates new samples from $\boldsymbol{x}_{\sigma}$ by reversing the forward SDE via the reverse SDE:
	\begin{equation}
	\mathrm{d} \boldsymbol{x}=\left[\mathbf{f}(\boldsymbol{x}, \sigma)-g(\sigma)^2 \nabla_{\boldsymbol{x}} \log p_\sigma(\boldsymbol{x})\right] \mathrm{d} \sigma+g(\sigma) \mathrm{d} \overline{\omega}_\sigma, 
	\label{eq:forward_sde} 
	\end{equation}
	where $\overline{\omega}_\sigma$ is also a standard Wiener process, and $p_\sigma(\boldsymbol{x})$ is the marginal probability density at $\sigma$. The only unknown term $\log p_\sigma(\boldsymbol{x})$ is estimated via score matching\cite{hyvarinen2005estimation}, in which a denoiser function   $\boldsymbol{D}_{\theta}(\boldsymbol{x}_{\sigma} ; \sigma)$ is learned, i.e., 
	\begin{equation}
	\mathbb{E}_{\boldsymbol{x}, \sigma} \|\boldsymbol{D}_{\theta}(\boldsymbol{x}_{\sigma} ; \sigma)-\boldsymbol{x}\|_2^2,  \nabla_{\boldsymbol{x}} \log p_\sigma(\boldsymbol{x})=\frac{\boldsymbol{D}_{\theta}(\boldsymbol{x}_{\sigma} ; \sigma)}{\sigma^2}.
	\label{eq:reverse_sde} 
	\end{equation}

	During training, samples $\boldsymbol{x}$ from the dataset are corrupted through the forward diffusion process. The denoiser function $\boldsymbol{D}_{\theta}(\boldsymbol{x}_{\sigma} ; \sigma)$ is then trained to reconstruct $\boldsymbol{x}$ from $\boldsymbol{x}_\sigma$. In the case of conditional generation, the conditional input is additionally taken as input of $\boldsymbol{D}_{\theta}$, i.e., $\boldsymbol{D}_{\theta}(\boldsymbol{x}_{\sigma} ; \sigma, \mathcal{C})$. During inference, leveraging the trained $\boldsymbol{D}_{\theta}$, new samples given the conditional input $\mathcal{C}$ can be synthesized by solving the reverse SDE using a numerical SDE solver.
 	
    \section{Methodology}
    \label{sec:Methodology}
 
     \subsection{Task Definition}
 	\label{sec:formulation}
    In this work, we aim at predicting 3D semantic point clouds from multi-frame 3D radar cubes. 
    Specifically, we are given consecutive $K$ radar spherical cubes denoted as $\mathbf{C}_p = \{\mathbf{C}_{p_{i}} \in \mathbb{R}^{R\times E \times A}, i=1,2, \cdots, K \}$, along with corresponding $\mathbf{SE(3)}$ poses $T = \{T_i = (R_i, t_i) \mid R \in \mathbf{S O}(3), t \in \mathbf{R}^3, i=1,2, \cdots, K \}$, where $R, E, A$ represent the range, elevation, and azimuth dimensions of the radar cube, respectively. The task is to recover 3D point cloud denoted as $\boldsymbol{L}=\{(\mathcal{P}_i, s_i) \mid \mathcal{P}_i \in \mathbb{R}^3, s_i \in \{1, 2,...,S\}\}_{i=1}^N$, where $s_i$ is the semantic label with $S$ potential classes in total, from large-scale but noisy radar cubes.

    \subsection{Data Pre-processing}
    \label{sec:preprocessing}
    As introduced in Sect.\ref{sec:radar_signal_processing}, radar cubes are inherently massive. However, the majority of their voxels correspond to sidelobe artifacts or free-space regions rather than valid targets. This motivates our approach to perform filtering and sparsifying on the radar cube, leveraging sparse operators for feature extraction to reduce computational overhead. Furthermore, we propose parallel frame accumulation, which utilizes GPU acceleration to enable real-time multi-frame non-coherent accumulation in the Cartesian coordinate, thereby enhancing the SNR of the radar cube and facilitating subsequent tasks.
    
    The detailed procedure is presented in Algorithm \ref{alg:radar_cube_preprocessing}, where $\mathbf{C_p}$ and $T$ are taken as input, and the algorithm outputs filtered radar Cartesian cube (RCC) denoted as $\mathbf{C}_{c} \in \mathbb{R}^{X\times{Y}\times{Z}}$, along with voxel index $\mathcal{V}_c \in \mathbb{N}^{M \times 3}$ and corresponding power intensity $\mathcal{I}_c \in \mathbb{R}^{M}$, where $M$ is the number of non-empty voxels in $\mathbf{C}_{c}$. 
    In this work, we select $K=5$, that is, when performing frame accumulation, we consider the current frame and the past 4 frames. First, INTENSITY\_FILT($\mathbf{C}_{p_{i}}, q_{th}$) is applied to each radar spherical cube $\mathbf{C}_{p_{i}}$ to preserve only  top-$q_{th}\%$ points. It is noteworthy that coordinate transformation in spherical coordinates is nontrivial, so SPHE\_TO\_CART($\mathcal{P}_{p_i}$) achieves spherical-to-Cartesian transformation, and  COORD\_TRANSFORM($\mathcal{P}_{c_i}, T_i, T_k$) align these points of past frames to the current ($K$-th) frame spatially via homogeneous transformations. After spatial alignment, VOXELIZE\_AND\_FOV\_FILT($\mathcal{P}, \mathcal{I}$) filters out the points in the past frames that exceed the FOV of the current frame and voxelizes the remaining points. Then, non-coherent accumulation is achieved by PARALLEL\_INDEX\_ACC($\mathbf{C}_{c}, \mathcal{V}, \mathcal{I}$), which sums the power intensity of each point to the corresponding voxel index in a sparse and parallel manner via GPU acceleration. Finally, intensity filering is applied again to get $M=X\times{Y}\times{Z}\times{q_{th}\%}$ voxels. We thereby unify the dense multi-frame radar cubes into a sparse representation given the voxel size and resolution of X-Y-Z dimension. The entire process is visually illustrated in Fig.\ref{pic:preprocess_visualization}.
    
    On the other hand, we employ traditional CFAR to each radar spherical cube to obtain multi-frame RPCs. RPCs are spatially aligned to the current frame in the same way as radar cubes, and the number of points are expanded to be the same with radar cubes where the expanded voxels are set to "free" class. The RCC and RPC are used together as inputs for the subsequent algorithms.

	 \begin{figure}
	\centering
	\includegraphics[width=1.0\linewidth]{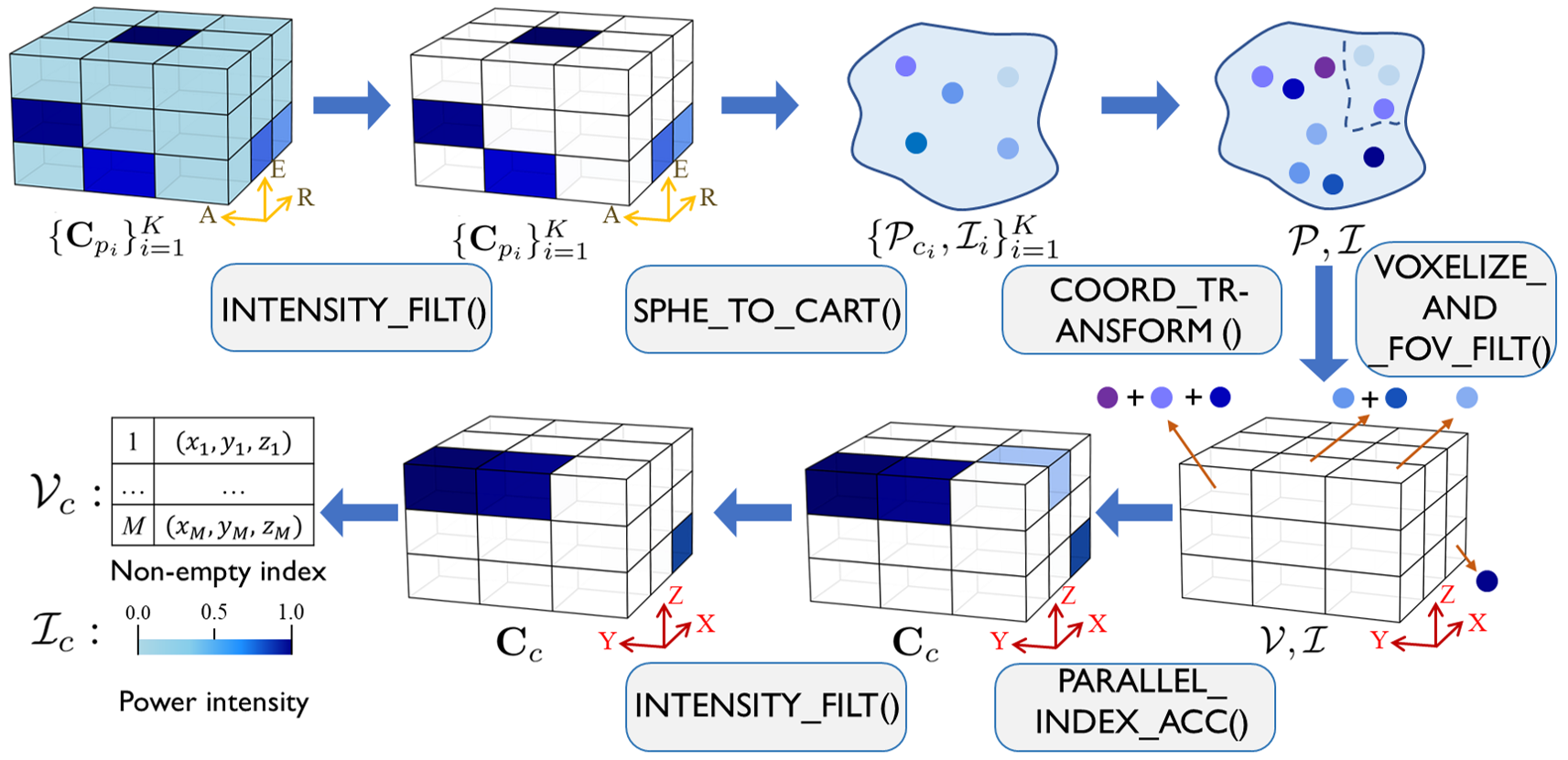}
	\caption{
		Illustration of the proposed radar cube pre-processing method which accumulates multi-frame radar spherical cubes into a intensity-filtered RCC represented in sparse format. 
	}
	\label{pic:preprocess_visualization}
	\end{figure}

	\begin{algorithm}[t]
		\label{alg:radar_cube_preprocessing}
		\DontPrintSemicolon
		\LinesNumbered
		\SetAlgoLined
		\caption{\textbf{Radar Cube Pre-processing}}
		\KwIn{~~radar spherical cubes $\mathbf{C}_{p}$, ego poses $T$}
		\KwOut{accumulated radar Cartesian cube $\mathbf{C}_{c}$, non-empty voxel index $\mathcal{V}_c$, power intensity $\mathcal{I}_c$}
		\Begin
		{
			$\mathcal{P}\leftarrow \{\}$\;
			$\mathcal{I}\leftarrow \{\}$\;
			
			/* \textbf{GPU data uploads} */ \; 
			
			\ForEach{$\mathbf{C}_{p_{i}} \in \mathbf{C}_{p}$}
			{
				$\mathbf{C}_{p_{i}}, \mathcal{P}_{p_i}, \mathcal{I}_i \leftarrow \text{INTENSITY\_FILT}(\mathbf{C}_{p_{i}}, q_{th})$\;
				$\mathcal{P}_{c_i} \leftarrow \text{SPHE\_TO\_CART}(\mathcal{P}_{p_i})$\;		
				\uIf {$i \neq K$} {
					$\mathcal{P}_{b_i} \leftarrow \text{COORD\_TRANSFORM}(\mathcal{P}_{c_i}, T_i, T_K)$\;
				} \Else {
					$\mathcal{P}_{b_i} \leftarrow \mathcal{P}_{c_i}$\;
				}

				$\mathcal{P}\leftarrow \mathcal{P}\cup\{\mathcal{P}_{b_i}\}$\;
				$\mathcal{I}\leftarrow \mathcal{I}\cup\{\mathcal{I}_i\}$\;
			}
			$\mathcal{V}, \mathcal{I} \leftarrow \text{VOXELIZE\_AND\_FOV\_FILT} (\mathcal{P}, \mathcal{I})$\;
			$\mathbf{C}_{c} \leftarrow \text{PARALLEL\_INDEX\_ACC}(\mathbf{C}_{c}, \mathcal{V}, \mathcal{I}) $\;
			$\mathbf{C}_{c}, \mathcal{V}_c, \mathcal{I}_c \leftarrow \text{INTENSITY\_FILT}(\mathbf{C}_{c}, q_{th})$\;
			\Return{$\mathbf{C}_{c}, \mathcal{V}_c, \mathcal{I}_c$}
		}
	\end{algorithm}
    
    \subsection{Stage \text{\uppercase\expandafter{\romannumeral1}}: Radar Cube Filtering for  coarse-grained structural and semantic mask prediction}
    \label{sec:stage1}
    After data pre-processing, the generated RCC still contains a large number of prominent sidelobe artifacts, making it difficult to identify target structures, hence resulting in a significant domain gap with LiDAR point cloud. To tackle this, we propose learning a coarse structure and suppressing sidelobe artifacts in Stage {\uppercase\expandafter{\romannumeral1} to bridge the modality gap. We first voxelize the LiDAR point cloud in the same 3D Cartesian cube as radar data, then apply gaussian blurring and greyscale dilation with the same kernel size to the binary (free or occupied) and semantic (with class label) version of LiDAR cubes, respectively. The former undergoes Gaussian blurring, with confidence levels that gradually decrease from the center of valid targets toward the margins, providing structural information, while the latter provides semantic information. Both are transformed in sparse representation and taken as the ground truth for Stage {\uppercase\expandafter{\romannumeral1}. As illustrated in Fig.\ref{pic:architecture}, gaussian blurring and greyscale dilation significantly enhance weak targets, such as wires. This effectively promoting the complete prediction of these weak targets, as discussed in Section \ref{sec:ablation}. 
    
    The problem is modeled as discriminative multi-task regression and classification. Concretely, with the common voxel index matrix $\mathcal{V}_c \in \mathbb{N}^{M \times 3}$, RCC $\mathcal{I}_{c} \in \mathbb{R}^{M \times 1}$ (filled with normalized power intensity) and RPC $\mathcal{I}_{p} \in \{0, 1\}^{M \times 1}$ (filled with free-or-occupied booleans) are concatenated as the input with shape $(M, 2)$. The confidence level of the structural label $y^{st}$ is normalized to [0,1], representing the probability that each voxel is non-empty, while the semantic label $y^{se}$ is expressed through one-hot encoding. The network $\boldsymbol{R}_\theta$ then predicts a structural mask $\hat{y}^{st}$ with shape $M \times 1$ and semantic mask $M \times S$ with shape $(M, S)$ through regression and classification heads, respectively. 
    
    During training, structural regression is trained with a binary cross-entropy (BCE) loss, which aims to minimize the probability distribution difference between the predicted mask and ground truth label. Semantic prediction is trained a weighted cross-entropy (WCE) loss that strikes a balance between different classes (e.g., "ground" class vs "pole" class in agricultural fields). Therefore, the overall loss function can be derived as:

	\begin{equation}
		\mathcal{L}_{stage1}= \mathcal{L}_{bce}(\hat{y}^{st}, {y}^{st}) + \mathcal{L}_{wce}(\hat{y}^{se}, {y}^{se}).
		\label{eq:stage1_loss}
	\end{equation}

    During inference, the output voxels with "free" class label are filtered out, with $L \textless M$ voxels left. Afterwards, we reduce the shape of the semantic mask to $L \times 1$ through argmax operation, then concatenate the structural and semantic mask to constitute a feature matrix denoted as $\mathcal{C}$ with shape $L \times (1+1)$, which serve as the conditional input of the fine-grained diffusion learning in Stage {\uppercase\expandafter{\romannumeral2}.

    \subsection{Stage \text{\uppercase\expandafter{\romannumeral2}}: Diffusion-based Fine-Grained 3D Semantic Point Cloud Generation}
    \label{sec:stage2}
    This section expounds on the second learning stage that leverages diffusion model for predicting fine-grained 3D LiDAR-like semantic point cloud from learned hybrid structural and semantic radar feature matrix $\mathcal{C}$. The whole process consists of diffusion training, diffusion inference, and consistency distillation. The proposed diffusion training and inference scheme follows the EDM\cite{karras2022elucidating} diffusion framework.
    
    \textbf{Diffusion Training} refers to training a neural network that reconstructs noise-corrupted semantic LiDAR point cloud conditioned on spatial-temporal aligned $\mathcal{C}$. It starts by perturbing LiDAR data sample $\boldsymbol{x} \sim p_{data}$ expressed in one-hot format through forward diffusion process $p(\boldsymbol{x}_{\sigma}\mid\boldsymbol{x}) \sim\mathcal{N}(\boldsymbol{x}, \sigma^2 \mathbf{I})$, where $\sigma \sim p_{noise}$ is the noise level sampled from a predefined noise schedule. Before forward diffusion, $\boldsymbol{x}$ is expanded to $L$ points where the newly added points are assigned the "free" class, then $\sigma$ is sampled with the same shape $L \times S$ as well.
    
     To recover $\boldsymbol{x}$ from $\boldsymbol{x}_{\sigma}$, a denoiser function $\boldsymbol{D}_{\theta}(\boldsymbol{x}_{\sigma} ; \sigma, \mathcal{C})$ is learned. Considering the noise scaling issue, we define $\boldsymbol{D}_{\theta}$ in the manner of EDM as:
	\begin{equation}
	\boldsymbol{D}_\theta(\boldsymbol{x}_{\sigma} ; \sigma, \mathcal{C})=c_{\text {skip }}(\sigma) \boldsymbol{x}_{\sigma}+c_{\text {out }}(\sigma) \boldsymbol{F}_\theta(c_{\text {in }}(\sigma) \boldsymbol{x}_{\sigma} ; \sigma, \mathcal{C} ),
	\label{eq:dtheta_definition}
	\end{equation}
	where $\boldsymbol{F}_\theta$ is a neural network to be trained, $c_{\text {skip }}$, $c_{\text {out}}$, and $c_{\text {in}}$ are scaling functions. We apply $\boldsymbol{x}$-prediction\cite{salimansprogressive} that directly estimates $\boldsymbol{x}$ by optimizing $\boldsymbol{D}_\theta$ with the following denoising objective:
	\begin{equation}
	\mathcal{L}_{stage2}=\mathbb{E}_{\sigma, \boldsymbol{x}, \mathcal{C}}\left[\lambda(\sigma) \cdot w_c  \left\|\boldsymbol{D}_\theta(\boldsymbol{x}_{\sigma} ; \sigma, \mathcal{C}) - \boldsymbol{x} \right\|_2^2\right],
	\end{equation}
	where $\lambda(\sigma)$ is a loss weight which equalizes the initial training loss over the entire $\sigma$ range\cite{karras2022elucidating}, $w_c$ is a class weight that deals with the class-imbalance issue. From this, $\boldsymbol{D}_{\theta}$ learns to reconstruct the noise-corrupted LiDAR data sample $\boldsymbol{x}_{\sigma}$ conditioned on learned radar feature matrix $\mathcal{C}$.
	

	\textbf{Diffusion Inference} involves the reverse diffusion process that predicts LiDAR data sample $\boldsymbol{x}_0$ from Gaussian noise $\boldsymbol{x}_{\sigma_{max}} \sim \mathcal{N}(\mathbf{0}, \sigma_{max}^2 \mathbf{I})$ under conditional input $\mathcal{C}$.
	This is achieved by solving the reverse SDE (Eq. \ref{eq:reverse_sde}) using a $2^{nd}$ Runge-Kutta solver, during which $\boldsymbol{D}_\theta$ needs to be iteratively sampled in $n$ discrete steps $\{\sigma\}_{i=1}^n$, where $\sigma_1 = \sigma_{min} < \sigma_2 < \cdots < \sigma_{n} = \sigma_{max}$.
	
	\textbf{Consistency Distillation} is incorporated in Sem-RaDiff for one-step inference. A consistency model\cite{song2023consistency} is trained in this stage to learn a consistency function $\boldsymbol{f}_\theta:(\boldsymbol{x}_{\sigma}, \sigma) \mapsto \boldsymbol{x}_{\sigma_{min}}$, where $\sigma_{min}$ is a small positive number that $\boldsymbol{x}_{\sigma_{min}}$ approximates  $\boldsymbol{x}_{0}$. Therefore, $\boldsymbol{f}_\theta$ points arbitrary pairs of $(\boldsymbol{x}_{\sigma}, \sigma)$ to the original sample $\boldsymbol{x}_0$ and we are able to directly predict LiDAR data sample from $\boldsymbol{x}_{\sigma_{max}}$ and $\mathcal{C}$ in only one network forward pass. In practice, $\boldsymbol{f}_\theta$ is distilled from a pre-trained diffusion model $\boldsymbol{D}_{\theta}$ via a simple MSE loss. For more comprehensive technical details, we refer the reader to \cite{song2023consistency}.
    
    After network prediction, the "free" voxels of the output $\hat{\boldsymbol{x}}$ are filtered out again, leaving $N \textless L$ points. Then, $\hat{\boldsymbol{x}}$ is converted to 3D semantic point cloud $\hat{\boldsymbol{L}}$ with shape $N \times 4$.
    The entire training and inference procedure of Sem-RaDiff (after data pre-processing) is described in Algorithm \ref{alg:training_algorithm},\ref{alg:inference_algorithm}. In addition, the attached video shows the dynamic visualization of the entire procedure.

    \begin{figure*}[t]
	\centering
	\includegraphics[width=0.95\linewidth]{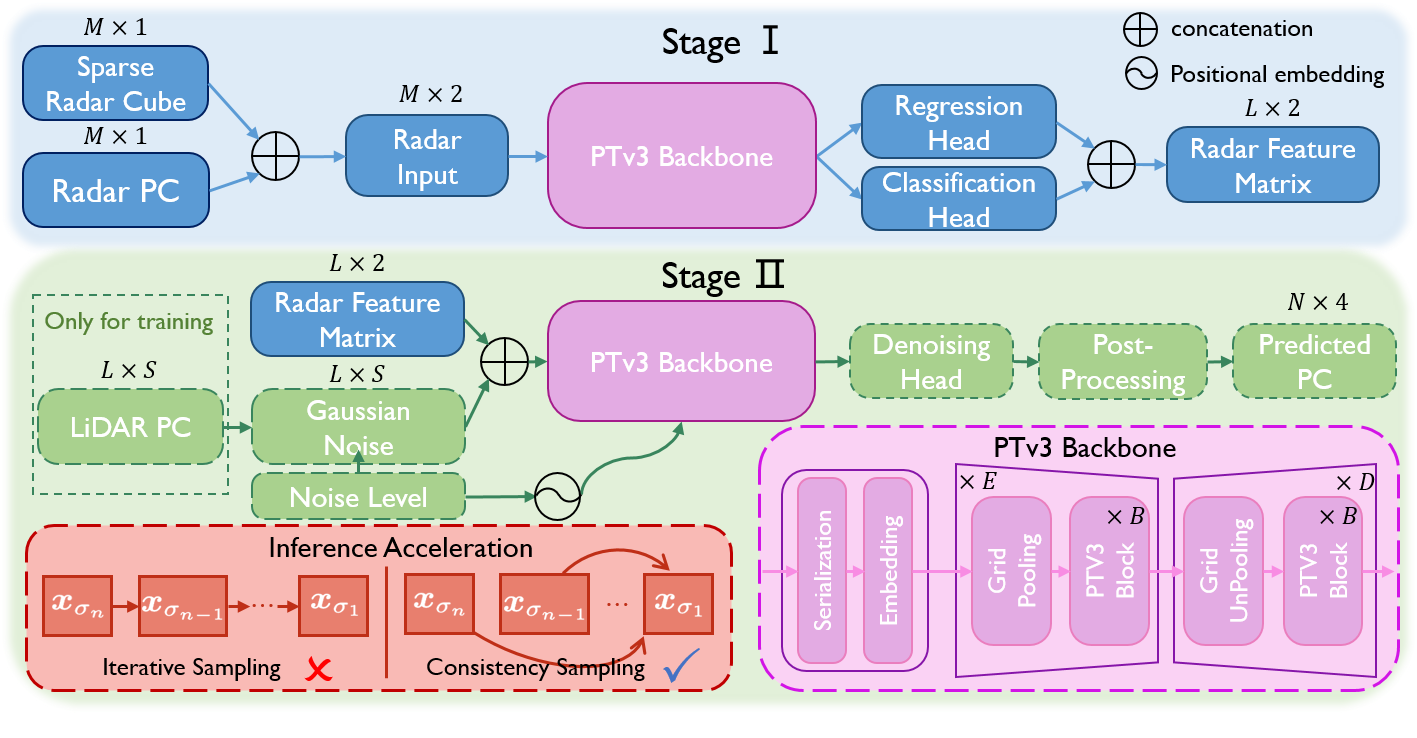}
	\caption{
		Overall architecture of the proposed sparse network for both Stage \uppercase\expandafter{\romannumeral1} and Stage \uppercase\expandafter{\romannumeral2}.
	}
	\label{pic:network_architecture}
	\vspace{-0.4cm}
	\end{figure*}

%
%
    
    \subsection{Model Architecture}
    \label{sec:model_architecrue}
	The model architecture of Sem-RaDiff is shown in Fig.\ref{pic:network_architecture}. We use Point Transformer V3 (PTv3)\cite{wu2024point} as the backbone for both learning stages. The PTv3 backbone is build on the U-Net framework\cite{ronneberger2015u}. It starts with point cloud serialization and embedding, reconstructing the originally unstructured point clouds (i.e., non-empty voxels) into a structured representation. Next is a sparse 3D U-Net\cite{ronneberger2015u} which consists of $E=5$ encoders and $D=4$ decoders. Each encoder/decoder comprises grid pooling/unpooling\cite{wu2022point} and PTv3 blocks, with each block containing a 3D sparse convolution layer and a multi-head attention layer as its core components. The specific configuration is the network is listed in Table \ref{tab:model_parameters}. In Stage \uppercase\expandafter{\romannumeral1}, both the structural and semantic head are a 3D sparse convolution layer with $1$ and $S$ output channels, respectively. In Stage \uppercase\expandafter{\romannumeral2}, conditional diffusion denoising is achieved by concatenating the $\mathcal{C}$ and $\mathbf{x}_{\sigma_{max}}$ to the network. We apply sinusoidal position embedding\cite{ho2020denoising} to the noise level $\sigma$ and then feed it to the network as an additional input. It is fused with the point feature in each PTv3 block by summation. The denoising head is a 3D sparse convolution layer with $S$ output channels. In terms of the noise distribution, scaling functions, loss weight,etc., we adopt the same settings as EDM\cite{karras2022elucidating}.

	\begin{algorithm}[t]
	\label{alg:training_algorithm}
	\DontPrintSemicolon
	\LinesNumbered
	\SetAlgoLined
	\caption{\textbf{Training Procedure of Sem-RaDiff}}
	{}
	\KwIn{radar inputs $\{\mathcal{I}_{c}, \mathcal{I}_{p}\}$, LiDAR point cloud $\boldsymbol{L}$}
	\KwOut{learned models $\boldsymbol{R}_\theta$, $\boldsymbol{f}_\theta$}
	/* \textbf{Stage \uppercase\expandafter{\romannumeral1}\textemdash training} */ \; 
	Voxelize LiDAR point cloud as $\boldsymbol{L}_c$. 
	
	Obtain the ground truth $\{y^{st}, y^{se}\}$ from $\boldsymbol{L}_c$ via Gaussian blurring and greyscale dilation.
	
	Train the network $\boldsymbol{R}_\theta$ via training loss $\mathcal{L}_{stage1}$.
	
	/* \textbf{Stage \uppercase\expandafter{\romannumeral2}\textemdash EDM training} */ \; 
	Obtain conditional input $\mathcal{C}$ from $\{\mathcal{I}_{c}, \mathcal{I}_{p}\}$ and $\boldsymbol{R}_\theta$.
	
	Expand $\boldsymbol{L}_c$ as a data sample $\boldsymbol{x}$.
	
	Corrupt $\boldsymbol{x}$ to $\boldsymbol{x}_\sigma$ by forward diffusion process.
	
	Train the network $\boldsymbol{D}_\theta$ via training loss $\mathcal{L}_{stage2}$.
	
	/* \textbf{Stage \uppercase\expandafter{\romannumeral2}\textemdash consistency distillation} */ \; 
	Train the network $\boldsymbol{f}_\theta$ by distilling $\boldsymbol{D}_\theta$ via MSE loss.
	
	\Return{$\boldsymbol{f}_\theta$}

	\end{algorithm}

	\begin{algorithm}[t]
	\label{alg:inference_algorithm}
	\DontPrintSemicolon
	\LinesNumbered
	\SetAlgoLined
	\caption{\textbf{Inference Procedure of Sem-RaDiff}}
	\KwIn{radar inputs $\{\mathcal{I}_{c}, \mathcal{I}_{p}\}$, learned models $\boldsymbol{R}_\theta$, $\boldsymbol{f}_\theta$}
	\KwOut{3D semantic point cloud prediction $\hat{\boldsymbol{L}}$}
	/* \textbf{Stage \uppercase\expandafter{\romannumeral1}\textemdash coarse mask prediction} */ \; 
	Predict the structural and semantic mask $\{\hat{y}^{st}, \hat{y}^{se}\}$ from $\boldsymbol{R}_\theta(\mathcal{I}_{c}, \mathcal{I}_{p})$.
	
	Filtering free voxels of $\{\hat{y}^{st}, \hat{y}^{se}\}$ and generate conditional input $\mathcal{C}$.
	
	/* \textbf{Stage \uppercase\expandafter{\romannumeral2}\textemdash fine point cloud prediction} */ \; 
	Sample Gaussian noise  $\boldsymbol{x}_{\sigma_{max}} \sim \mathcal{N}(\mathbf{0}, \sigma_{max}^2 \mathbf{I})$.
	
	Predict 3D semantic voxels $\hat{\boldsymbol{x}}$ from $\boldsymbol{f}_\theta(\boldsymbol{x}_{\sigma_{max}}, \mathcal{C})$.
	
	Converting voxels grids $\hat{\boldsymbol{x}}$ to point clouds $\hat{\boldsymbol{L}}$.
	
	\Return{$\hat{\boldsymbol{L}}$}
	
	\end{algorithm}

    \section{Results}
    \label{sec:results}
    \subsection{Dataset Preparation and Evaluation Metrics}
    
    In view of the scarcity of onboard mmWave radar datasets in agricultural scenarios, we build a dataset in agricultural fields using a commercial agricultural drone platform. As shown in Fig.\ref{pic:Background}, the drone carries a mmWave radar, a vertically placed LiDAR, and a front-view RGB camera. The radar and the LiDAR are spatially calibrated and temporally synchronized to jointly record data at a frequency of 5Hz. The FOV of both the radar and the LiDAR are cropped to [4m, 40m] in range, ±45° in elevation, and ±25° in azimuth. 
    Accordingly, the shape of radar spherical cubes is trimmed to $(R = 166) \times (E = 94) \times (A = 177)$. After data pre-processing, The shape of the common 3D Cartesian cube is $(X = 150) \times (Y = 150) \times (Z = 100)$ which corresponds to [4m, 40m], [-20m, 20m], [-20m, 10m] in X, Y, and Z dimension, respectively.
    The RGB camera is used to capture the scenes. In addition, an real-time kinematic positioning (RTK) module provides accurate position and attitude information.
    
    Our dataset includes 8 sequences, each corresponding to a distinct flight trajectory. The first 6 sequences were recorded in one agricultural field, with a total of 12728 frames of data serving as the training set. The latter 2 sequences are recorded in another agricultural field, with a total of 2587 frames of data serving as the test set. The LiDAR point clouds are labeled with 5 classes ("free", "ground", "tree", "pole", "wire"). 
    
    As for evaluation metrics, we report the Intersection over Union (IoU), precision, recall, and Chamfer Distance (CD) as the geometric metric, as well as the IoU of each class and the mean IoU as the semantic metric. Besides, the "free" class is excluded from the evaluation. These evaluation metrics provide a comprehensive quantification of:
    (1) false alarm and missed detection rates of the predicted radar point clouds $\mathcal{P}_R$ relative to the LiDAR ground truth $\mathcal{P}_L$,
    (2) geometric similarity between  $\mathcal{P}_R$ and $\mathcal{P}_L$, and
    (3) semantic prediction accuracy of $\mathcal{P}_R$. The metrics are described in Table \ref{tab:metric}.
 
	\begin{table}[t]
	
	\caption{Definition of Evaluation Metrics}
	\label{tab:metric}
	\centering
	\renewcommand{\arraystretch}{1.6}
	\setlength\tabcolsep{2.5pt}
	\resizebox{1.0\linewidth}{!}{%
		\fontsize{6}{6}\selectfont
		\begin{threeparttable}
		\begin{tabular}{l|l}
			\toprule
		 	 $TP$ & $\sum_{p_i \in \mathcal{P}_{{R}}}  \mathbb{I}\left(\min _{q_j \in \mathcal{P}_{{L}}}\left\|p_i-q_j\right\|_2 \leq \tau\right)$ \\
			 $FP$ & $\sum_{p_i \in \mathcal{P}_{{R}}}  \mathbb{I}\left(\min _{q_j \in \mathcal{P}_{{L}}}\left\|p_i-q_j\right\|_2 > \tau\right)$ \\
			 $FN$ & $\sum_{q_j \in \mathcal{P}_{{L}}}  \mathbb{I}\left(\min _{p_i \in \mathcal{P}_{{R}}}\left\|q_j - p_i\right\|_2 > \tau\right)$ \\
			 $D1$ &  $\frac{1}{TP+FP} \sum_{p_i \in \mathcal{P}_{{R}}}  \min _{q_j \in \mathcal{P}_{{L}}}\left\|p_i-q_j\right\|_2$ \\
			 $D2$ & $\frac{1}{TP+FN} \sum_{q_j \in \mathcal{P}_{{L}}}  \min _{p_i \in \mathcal{P}_{{R}}}\left\|q_j-p_i\right\|_2$ \\
			 \textbf{CD} &  $(D1 + D2) / 2$ \\
			 \textbf{Precision} & $TP / (TP + FP)$\\ 
			 \textbf{Recall} & $TP / (TP + FN)$\\ 
			 \textbf{IoU} & $TP / (TP + FP + FN)$\\ 
			 \textbf{mIoU} & $\frac{1}{S} \sum_{i=1}^{S} (TP_i / (TP_i + FP_i + FN_i))$ \\
			\bottomrule
		\end{tabular}
        \begin{tablenotes}
			\fontsize{6}{6}\selectfont
			\item[1] $\tau$ denotes the range threshold.
			\item[2] CD is short for Chamfer distance.
			\item[3] $TP_i, FP_i, FN_i$ denotes $TP, FP, FN$ for the $i$-th class, respectively.
		\end{tablenotes}
		\end{threeparttable}
		
		}
	\vspace{-0.1cm}
	\end{table}

\begin{table}[t]
	\caption{Model configuration of Sem-RaDiff}
	\label{tab:model_parameters}
	\centering
	\renewcommand{\arraystretch}{1.6}
	\setlength\tabcolsep{2.0pt}
	\resizebox{1.0\linewidth}{!}{%
		\begin{threeparttable}
		\begin{tabular}{lcc}
			\toprule[1.5pt]
			\text { Parameter } & \text { \textbf{Ours-M} } & \text { \textbf{Ours-L} } \\
			\midrule[1pt]
			\text { Encoder Depth } & \text { [2, 2, 2, 6, 2] } & \text { [3, 3, 3, 9, 3] } \\
			\text { Encoder Channels } & \text { [32, 64, 128, 256, 256] } & \text { [32, 64, 128, 256, 512] } \\
			\text { Decoder Depth } & \text { [2, 2, 2, 2] } & \text { [3, 3, 3, 3] } \\
			\text { Decoder Channels} & \text { [64, 64, 128, 256] } & \text { [64, 64, 128, 256] } \\
			\midrule[1pt]
			\text {Attention Patch Size } & \text { 1024 } & \text { 2048 } \\
			\text { Encoder Heads } & \text { [2, 4, 8, 16, 16] } & \text { [4, 8, 16, 32, 32] } \\
			\text { Decoder Heads } & \text { [4, 4, 8, 16] } & \text { [8, 8, 16, 32] } \\
			\bottomrule[1.5pt]
		\end{tabular}
        \begin{tablenotes}
			\item[1] We refer readers to PTv3\cite{wu2024point} for details about the above parameters.
		\end{tablenotes}

		\end{threeparttable}
	}
			\vspace{-0.35cm}
\end{table}

\begin{table}
	\setlength{\tabcolsep}{0.0035\linewidth}
	\newcommand{\classfreq}[1]{{~\tiny(\semkitfreq{#1}\%)}}  %
	\centering
	\caption{Comparison of input Modality and Model Efficiency }
	\label{tab:method_property}
	\resizebox{1.0\linewidth}{!}{
		\fontsize{5.5}{5.5}\selectfont
		\begin{threeparttable}
		\begin{tabular}{l|c| c | c c c}
			
			\toprule
			Method 
			& Modality
			& \begin{tabular}[c]{@{}c@{}}Params. \\(MB)\end{tabular}
			& \begin{tabular}[c]{@{}c@{}}Memory \\(GB)\end{tabular}
			& \begin{tabular}[c]{@{}c@{}}Comp. \\Eff.\end{tabular}
			& Num. S.
			\\
			\midrule
			RadarHD$^\star$ & C & 106.4 & 2.65 & 141.53 & \textbf{1} \\
			Sun$^\star$ & C & 133.2 & 2.83 & 153.27 & \textbf{1} \\
			\midrule
			Radar-Diffusion$^\star$ & C & 24.1 & 4.73 & 1295.48 & 40 \\
			Luan$^\star$ & P & 24.1 & 4.73 & 1295.48 & 40 \\
			\midrule
			\textbf{Ours-M (EDM)} & C\&P & 30.8 & \textbf{1.92}  & \textbf{68.89} & 40 \\
			\textbf{Ours-M (CD)}  & C\&P & 30.8 & \textbf{1.92}  & \textbf{68.89} & \textbf{1} \\
			\textbf{Ours-L (EDM)} & C\&P & 82.7 & 2.64           & 168.71         & 40 \\
			\bottomrule
		\end{tabular}
        \begin{tablenotes}
		\item[1] C and P denotes RCC and RPC, respectively.
		\item[2] Comp. Eff. indicates the computational efficiency measured by Giga Floating-point Operations Per Second (GFLOPS).
		\item[3] Num. S. indicates the number of forward evaluations required for one inference stage.
		\end{tablenotes}
		
	\end{threeparttable}
}
\vspace{-0.45cm}
\end{table}

\begin{figure*}[t]
	\centering
	\includegraphics[width=0.85\linewidth]{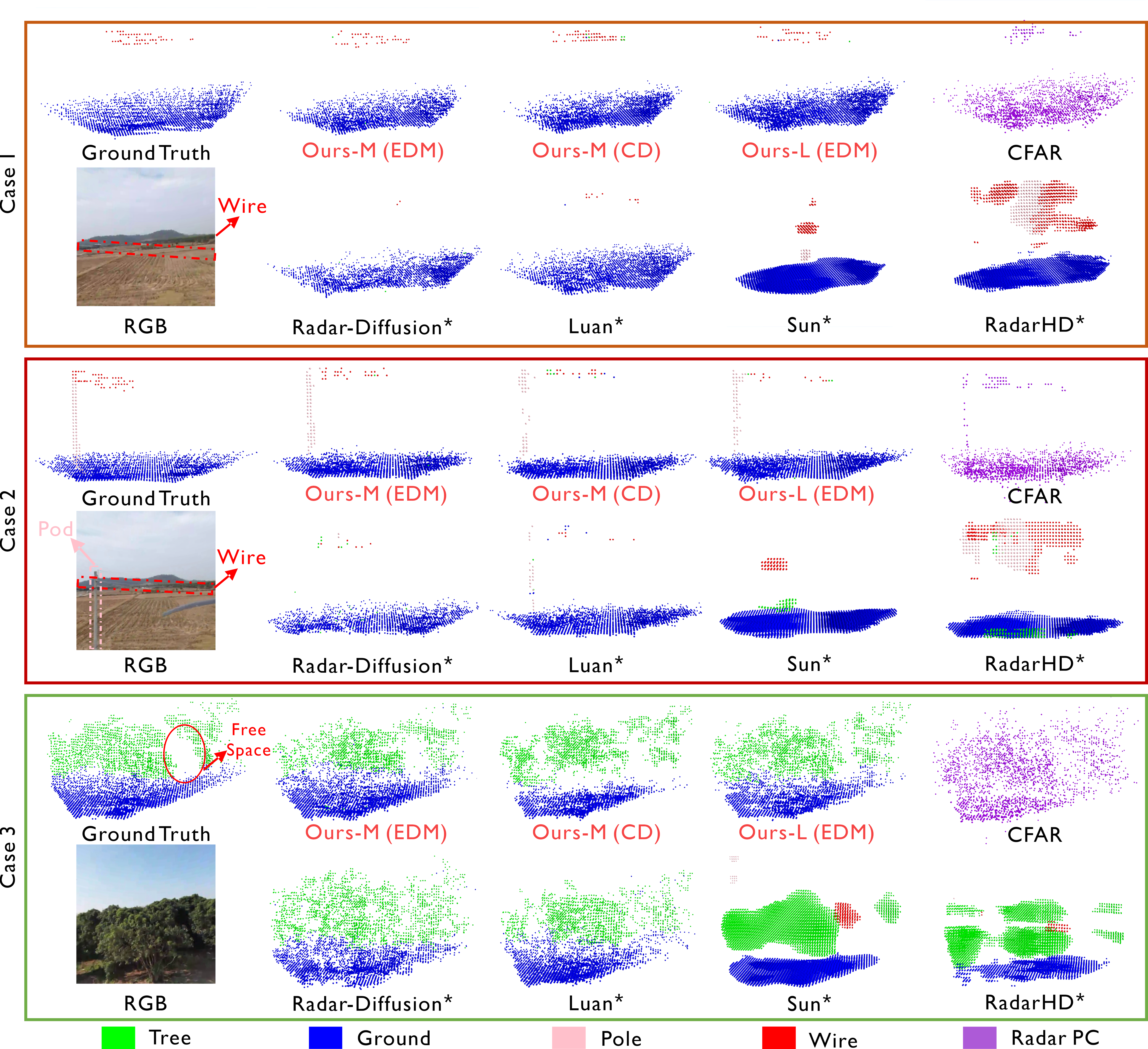}
	\caption{
		Qualitative results of single-frame 3D radar point cloud reconstruction and semantic prediction on the self-built dataset (test set). 
	}
	\label{pic:single_frame_visualization}
		\vspace{-0.2cm}
\end{figure*}

\begin{figure*}[t]
	\centering
	\includegraphics[width=0.85\linewidth]{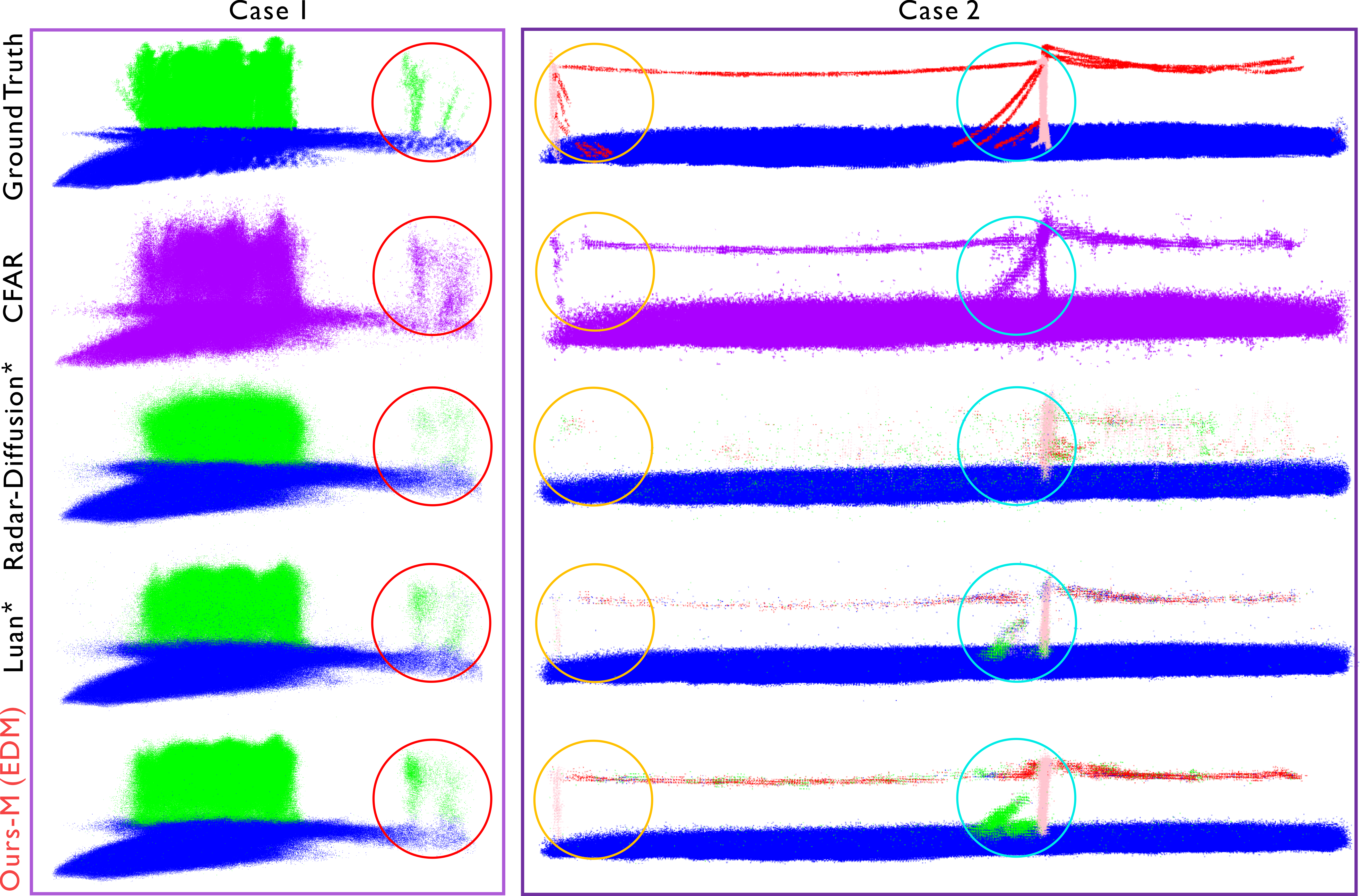}
	\caption{
		Qualitative results of multi-frame accumulated 3D radar point cloud reconstruction and semantic prediction on the self-built dataset (test set). . 
	}
		\vspace{-0.1cm}
	\label{pic:multiframe_visualization}

\end{figure*}
 	\label{sec:dataset}
     \subsection{Baselines and Training Details}
 	\label{sec:training_details}
    We benchmark the performance of Sem-RaDiff against 5 baseline methods, including a CFAR radar target detector, discriminative learning-based methods RadarHD\cite{prabhakara2023high} and the approach by Sun et al.\cite{sun2025automatic}, and diffusion-based methods Radar-Diffusion\cite{zhang2024towards}, as well as another diffusion-based scheme by Luan et al.\cite{luan2024diffusion}. Technical details for all learnning-based methods are listed here:
	\begin{itemize}
		\item Ours-M (EDM): The default version which is trained on a middle-size (with \textbf{30.8M} parameters) model and inferred through iterative sampling. Both Stage \uppercase\expandafter{\romannumeral1} and Stage \uppercase\expandafter{\romannumeral2} models are trained on 1 H800 GPU with a batch size of 12 and a learning rate of 1e-4. The former is trained for 20 epochs using the Adam optimizer, while the latter is trained for 100 epochs using the RAdam\cite{liuvariance} optimizer. Detailed model configuration is shown in Table \ref{tab:model_parameters}.
		\item Ours-M (CD): The one-step inference version which uses the same middle-size model and inferred through consistency sampling. The consistency model is distilled from the middle-size model after 50 epochs of training.
		\item Ours-L (EDM): The large version which is trained on a large-size (with \textbf{82.7M} parameters) model and inferred through iterative sampling. The detailed model configuration is shown in Table \ref{tab:model_parameters}.
		\item RadarHD$^\star$~\cite{prabhakara2023high}: The 3D version of RadarHD. In the original approach, 2D radar range-azimuth heatmaps (RAHs) are used as the input to the network, which predicts BEV point clouds without considering semantic predictions. In our implementation, the network architecture is modified to a 3D U-Net (based on Unetr++\cite{shaker2024unetr++}, with \textbf{100M} parameters), taking radar spherical cubes as input and fine-grained LiDAR semantic point clouds as ground truth. The model is  trained on our dataset from scratch for 30 epochs with supervision from a combination of WCE and class-wise affinity loss\cite{cao2022monoscene}. 
		\item Sun$^\star$~\cite{sun2025automatic}: Details are the same as the original approach except that we adopt the same loss as RadarHD$^\star$, where the class-wise affinity loss is used instead of the soft Dice loss. This change is made as we observe an obvious improvement in semantic prediction with the class-wise affinity loss. The model is trained for 30 epochs on our dataset from scratch.
		\item Radar-Diffusion$^\star$~\cite{zhang2024towards}: The 3D version of our previous Radar-Diffusion. We take radar spherical cubes instead of RAHs as the conditional input and fine-grained LiDAR semantic point clouds as ground truth. We also replace the original 2D U-Net to 3D Diff-UNet\cite{xing2023diff} with \textbf{24.1M} parameters) that is designed specifically for diffusion-based volumetric segmentation\textbf{}. The model is trained for 100 epochs on our dataset from scratch.
		\item Luan$^\star$~\cite{luan2024diffusion}: The 3D version of the scheme by Luan et al.. The original approach takes RPC as the conditional input. It adopts BEV representation and predicts 3D point clouds by embedding height information in the channel dimension. This strategy is not suitable for our dataset as it fails to handle multi-layer structures, such as wires above ground. We thereby uses Diff-UNet with the same model and training configuration as Radar-Diffusion$^\star$ with 3D RPCs as the conditional input. 
	
	\end{itemize}    
	
	Comparison of input modality and model efficiency between Sem-RaDiff and baseline methods are shown in Table \ref{tab:method_property}. As can be seen, only the proposed Sem-RaDiff utilizes both RCC and RPC for network input. Moreover, Ours-M incurs minimal computational overhead, requiring 51.3\% fewer GFlops and 27.5\% less GPU memory consumption than the second-best method (RadarHD$^\star$). Furthermore, after consistency distillation, enables one-step inference like discriminative  methods. The qualitative and quantitative results of learning-based methods in Subsection \ref{sec:benchmark} are obtained via half-precision inference.

\begin{table*}
	\setlength{\tabcolsep}{0.0035\linewidth}
	\centering
	\caption{
		Quantitative results on the self-built dataset (test set) with different range thresholds
	}
	\label{tab:quantative_0.5m}
	\resizebox{1.0\linewidth}{!}{
		\begin{threeparttable}
			\begin{tabular}{l| c c c c c c c c c c | c c c c c c c}
				
				\toprule[1.5pt]
				\multirow{2}{*}{Method}
				& \multicolumn{2}{c}{mIoU (\%)}
				& \multicolumn{2}{c}{\textcolor{ground}{$\blacksquare$} Ground (\%)}
				& \multicolumn{2}{c}{\textcolor{tree}{$\blacksquare$} Tree (\%)}
				& \multicolumn{2}{c}{\textcolor{pole}{$\blacksquare$} Pole (\%)}
				& \multicolumn{2}{c|}{\textcolor{wire}{$\blacksquare$} Wire (\%)}
				& \begin{tabular}[c]{@{}c@{}}Chamfer \\ Distance (m)\end{tabular}
				& \multicolumn{2}{c}{IoU (\%)}
				& \multicolumn{2}{c}{Precision (\%)}
				& \multicolumn{2}{c}{Recall (\%)} \\
				
				& \multicolumn{1}{c}{0.25m}
				& \multicolumn{1}{c}{0.5m}
				& \multicolumn{1}{c}{0.25m}
				& \multicolumn{1}{c}{0.5m}
				& \multicolumn{1}{c}{0.25m}
				& \multicolumn{1}{c}{0.5m}
				& \multicolumn{1}{c}{0.25m}
				& \multicolumn{1}{c}{0.5m}
				& \multicolumn{1}{c}{0.25m}
				& \multicolumn{1}{c|}{0.5m}
				&
				& \multicolumn{1}{c}{0.25m}
				& \multicolumn{1}{c}{0.5m}
				& \multicolumn{1}{c}{0.25m}
				& \multicolumn{1}{c}{0.5m}
				& \multicolumn{1}{c}{0.25m}
				& \multicolumn{1}{c}{0.5m}		
				\\
				\midrule
				CFAR & - & -  & - & - & -  & - & - & -  & - & -  &  0.80 & -  & - &  37.40 & 75.84 & \lastnum{21.22} & \lastnum{60.86}\\
				\midrule
				RadarHD$^\star$ & 12.60  & 31.57 &  22.80 & 61.68 & \secondnum{23.03} & 50.92 & 3.50 &  8.22 & 1.16 & 5.47 &  1.52 & 22.70 & 58.61  & 29.37 & 70.27 & \secondnum{49.35} & 77.31\\
				Sun$^\star$ & 11.12 & 27.74 & 20.48 & 56.59 & \thirdnum{22.38}  & 49.97 & \lastnum{1.40} & \lastnum{3.53} & 0.21 & 0.88 &  \lastnum{1.64} & 20.81  & 54.25   & \lastnum{25.69} &  \lastnum{63.82} & \topnum{50.52} & \thirdnum{77.68}\\
				\midrule
				Radar-Diffusion$^\star$ & \lastnum{8.29} & \lastnum{24.10} & \lastnum{15.72} & \lastnum{54.71} & \lastnum{12.33} & \lastnum{30.70} & 5.03 & 10.65 & \lastnum{0.08} & \lastnum{0.32}  &  0.94 & \lastnum{15.10}  & \lastnum{49.69}  &  29.66 &  72.9 & 24.57 & 62.67\\
				Luan$^\star$ & 14.94  & 37.39& \secondnum{25.05} & \secondnum{71.58} & 19.97 & 48.67 & 13.01 & 23.91 & 1.74 & 5.41 & \secondnum{0.67} & 24.18 &  \secondnum{67.23} & \thirdnum{41.99} &   \thirdnum{83.31} & 36.53 & \secondnum{78.87}\\
				\midrule
				\textbf{Ours-M (EDM)} & \thirdnum{18.20} & \secondnum{44.90} & \thirdnum{24.60} & \thirdnum{69.14} & 22.26 & \secondnum{54.31} & \thirdnum{19.51}  & \thirdnum{37.24} & \topnum{6.45} & \topnum{18.89} & \thirdnum{0.69} & \thirdnum{24.29} & \thirdnum{66.13}  &  \secondnum{42.09} &  \secondnum{84.48} & 35.87 &  75.33\\
				\textbf{Ours-M (CD)} & \topnum{19.59} & \thirdnum{44.50} & 22.94 & 60.38 & \topnum{25.33} & \topnum{56.10} & \topnum{24.61} & \topnum{46.91} & \thirdnum{5.48} & \thirdnum{14.59} &  0.79 & \secondnum{24.84} & 62.16 &  41.98  &  83.29 & 35.99 & 68.89\\
				\textbf{Ours-L (EDM)} & \secondnum{18.98} & \topnum{45.98} & \topnum{28.54}  & \topnum{75.76} &  21.59  & \thirdnum{52.76} & \secondnum{19.96} & \secondnum{37.70} & \secondnum{5.80} & \secondnum{17.72} &  \topnum{0.63} & \topnum{27.45} &  \topnum{72.28}   &  \topnum{42.37}  &  \topnum{84.70} & \thirdnum{44.76} & \topnum{83.20}\\
				\bottomrule[1.5pt]
			\end{tabular}
			\begin{tablenotes}
				\normalsize
				\item[1] The best, second best, third best, and worst results are marked \topnum{red}, \secondnum{brown}, \thirdnum{pink}, and \lastnum{blue}, respectively.
			\end{tablenotes}
		\end{threeparttable}
	}
	\vspace{-0.35cm}
\end{table*}

    \subsection{Benchmark Comparison}
    \label{sec:benchmark}
    \subsubsection{Qualitative Comparison}
    \label{sec:qualitative}
    Fig.\ref{pic:single_frame_visualization} visually demonstrates single-frame examples of qualitative results from different methods. Compared to baseline methods, Sem-RaDiff provides denser and more accurate 3D structural and semantic prediction, especially for fine targets such as poles and wires. Specifically, discriminative methods (Sun$^\star$ and RadarHD$^\star$) fail to generate accurate LiDAR-like structures and produce a large number of artifacts. One-stage diffusion model-based methods (Radar-Diffusion$^\star$ and Luan$^\star$) can predict ground and tree classes more accurately than discriminative methods, but struggle with pole and wire classes, leading to missed detections. Specifically, Radar-Diffusion$^\star$, which uses only large-scale radar spherical cubes as conditional input, barely detected wires and poles in case 1 and case 2. In contrast, Sem-RaDiff precisely predicts both the structural and semantic features of targets under all configurations. In case 1, our method more completely generates both overhead wires compared to CFAR and other learning-based methods. In case 2, our method manages to detect wires and compete the pole. In case 3, our method predicts tree shapes and free space between trees more accurately than Radar-Diffusion$^\star$ and Luan$^\star$.
    Notably, even ours-M (CD) which features one-step inference outperforms Radar-Diffusion$^\star$ and Luan$^\star$ which rely on iterative sampling in predicting poles and wires. This is attributed to our proposed radar cube preprocessing and Stage \uppercase\expandafter{\romannumeral1}, which enhances SNR and suppresses substantial sidelobe artifacts, respectively.
    
     We further compare our method with the baseline methods in multi-frame cases. As shown in Fig. \ref{pic:multiframe_visualization}, in case 1, ours-M (EDM) best distinguishes two tiny trees (marked with red circle) and correctly predicts their class. In contrast, Luan$^\star$, while achieving comparable structural prediction results, exhibits significantly more errors in semantic prediction, misclassifying many tree-class point clouds as ground class. In case 2, our method most completely recovers the structure of the left-side pole marked with orange circle and overhead wires, with minimal misclassification. Our method also best reconstructs the structure of the slanting wires marked with blue circle, although all methods misclassifies them as trees. We attribute this to the training set containing only horizontal overhead wires without examples of slanting wires. Additionally, similar to baseline methods, ours misses the detection of the slanting wires in the orange circle. This occurs because our model tends to rely on RPC, which fails to detect these slanting wires. We provide detailed analysis of this issue in Section \ref{sec:ablation}.
 
 \begin{table*}
 	\setlength{\tabcolsep}{0.0035\linewidth}
 	\centering
 	\caption{
 		Ablation studies on key modules and input modalities of Sem-RaDiff
 	}
 	\label{tab:ablation}
 	\resizebox{1.0\linewidth}{!}{
 		\fontsize{4.5}{4.5}\selectfont
 		\begin{threeparttable}
 			\begin{tabular}{c|c c c c c c c c c c| c c c c c c c }
 				\toprule
 				\multirow{2}{*}{{Method}}
 				& \multicolumn{2}{c}{{mIoU (\%)}}
 				& \multicolumn{2}{c}{\textcolor{ground}{$\blacksquare$} Ground (\%)}
 				& \multicolumn{2}{c}{\textcolor{tree}{$\blacksquare$} Tree (\%)}
 				& \multicolumn{2}{c}{\textcolor{pole}{$\blacksquare$} Pole (\%)}
 				& \multicolumn{2}{c|}{\textcolor{wire}{$\blacksquare$} Wire (\%)}
 				& \begin{tabular}[c]{@{}c@{}}Chamfer \\ Distance (m)\end{tabular}
 				& \multicolumn{2}{c}{{IoU (\%)}}
 				& \multicolumn{2}{c}{{Precision (\%)}}
 				& \multicolumn{2}{c}{{Recall (\%)}}
 				\\
 				
 				& \multicolumn{1}{c}{0.25m}
 				& \multicolumn{1}{c}{0.5m}
 				& \multicolumn{1}{c}{0.25m}
 				& \multicolumn{1}{c}{0.5m}
 				& \multicolumn{1}{c}{0.25m}
 				& \multicolumn{1}{c}{0.5m}
 				& \multicolumn{1}{c}{0.25m}
 				& \multicolumn{1}{c}{0.5m}
 				& \multicolumn{1}{c}{0.25m}
 				& \multicolumn{1}{c|}{0.5m}
 				&
 				& \multicolumn{1}{c}{0.25m}
 				& \multicolumn{1}{c}{0.5m}
 				& \multicolumn{1}{c}{0.25m}
 				& \multicolumn{1}{c}{0.5m}
 				& \multicolumn{1}{c}{0.25m}
 				& \multicolumn{1}{c}{0.5m}
 				\\
 				\midrule

 				\begin{tabular}[c]{@{}c@{}}w/o SNR \\ Enhancing\end{tabular} & 17.39 & 43.09 & 24.23 & 68.49 & 20.24 & 50.24 & \textbf{22.51} & \textbf{45.20} & 2.86 & 8.44 & 0.70  & 23.64 &  65.08 & 41.35 & 84.01  & 34.79 & 73.66\\
  				\begin{tabular}[c]{@{}c@{}}w/o RCC \\ Filtering\end{tabular}  & 16.62 & 40.47 & 24.17 & 69.09 & 20.59 & 47.69 & 18.77 & 37.53 & 2.94 & 7.59 & 0.693  & 23.57 &  64.45 & 44.13 & \textbf{87.03}  & 33.18 & 72.13 \\
 				RCC only & \underline{11.97} & \underline{33.24} & \underline{19.36} & 64.40 & \underline{17.09} & \underline{43.98} & \underline{10.74} & \underline{22.28} & \underline{0.70} & \underline{2.30} & \underline{0.79}  & \underline{19.03} &  60.18 & \underline{33.03} & \underline{77.84}  & 32.80 & 74.70\\
 				RPC only & 16.85 & 41.74 & 23.27 & \underline{63.40} & 19.00 & 46.32 & 20.13 & 42.31 & 4.98 & 14.95 & 0.76  & 22.49 &  \underline{59.90} & \textbf{44.54} & 85.52  & \underline{30.51} & \underline{66.65}\\
 				\midrule
 				\begin{tabular}[l]{@{}l@{}} Ours-M \\ (EDM) \end{tabular} & \textbf{18.20} & \textbf{44.90} & \textbf{24.60} & \textbf{69.14} & \textbf{22.26} & \textbf{54.31} & 19.51 & 37.24 & \textbf{6.45} & \textbf{18.89} & \textbf{0.691}  & \textbf{24.29} &  \textbf{66.13} & 42.08 & 84.48  & \textbf{35.87} & \textbf{75.33} \\
 				\bottomrule
 			\end{tabular}
 			\begin{tablenotes}
 				\fontsize{5}{5}\selectfont
 				\item[1] The best and worst results are highlighted with \textbf{bold} and \underline{underline}, respectively.
 			\end{tablenotes}
 		\end{threeparttable}
 	}
 \vspace{-0.3cm}
 \end{table*}

 \begin{table}
 	\setlength{\tabcolsep}{0.0035\linewidth}
 	\centering
 	\caption{
 		Time Efficiency of Sem-RaDiff
 	}
 	\label{tab:model_efficiency}
 	\resizebox{0.85\linewidth}{!}{
 		\fontsize{4.0}{4.0}\selectfont
 		\begin{threeparttable}
 			\begin{tabular}{l|c|ccc|c|c}
 				\toprule
 				& \multicolumn{1}{c|}{\begin{tabular}[c]{@{}c@{}}PFA-CPU \\ (ms)\end{tabular}} & \multicolumn{1}{c}{\begin{tabular}[c]{@{}c@{}}PFA-GPU \\ (ms)\end{tabular}} & \begin{tabular}[c]{@{}c@{}}Stage \uppercase\expandafter{\romannumeral1} \\ (ms)\end{tabular} & \begin{tabular}[c]{@{}c@{}}Stage \uppercase\expandafter{\romannumeral2} \\ (ms)\end{tabular} & \begin{tabular}[c]{@{}c@{}}Total \\ (ms)\end{tabular}  & \begin{tabular}[c]{@{}c@{}}FPS \\ (Hz)\end{tabular}   \\
 				\midrule
 				Ours-M        & \multirow{2}{*}{162.6}      & \multirow{2}{*}{4.1}        & 44.5  & 24.7  & 73.3  & 13.6 \\
 				Ours-L        &                              &                              & 89.8  & 43.7  & 137.6 & 7.3 \\
 				\bottomrule
 			\end{tabular}
 			\begin{tablenotes}
 				\fontsize{3.5}{3.5}\selectfont
 				\item[1] PFA is short for parallel frame accumulation.
 			\end{tablenotes}
 		\end{threeparttable}
 	}
 \vspace{-0.6cm}
 \end{table}
 
    \subsubsection{Quantitative Comparison}
    \label{sec:quantitative}
    Quantitative results of our method and baseline methods on the test set of the self-built dataset is presented in Table \ref{tab:quantative_0.5m} which contains both semantic metrics (IoU of each class and mIoU) and geometric metrics (Chamfer distance, IoU, Precision and Recall) with difference range thresholds ($\sigma$=0.25m/0.5m). As shown in Table \ref{tab:quantative_0.5m}, our method achieves top-three rankings in most metrics across all three configurations. In terms of semantic metrics, our method demonstrates significant leads in both mIoU and per-class IoU. Particularly for challenging classes such as pole and wire, even Ours-M (CD) surpasses the best-performing baseline method (Luan*) by about 1~2 times in IoU. 
    
    In terms of geometric metrics, Ours-L (EDM) achieves the best performance compared to baseline methods, attaining the highest scores across all metrics except for the Recall metric at $\sigma$=0.25m. While RadarHD$^\star$ and Sun$^\star$ achieve higher scores on this metric, their performance on all other metrics is significantly inferior to our method. This indicates that the point clouds they generate are considerably less precise than those produced by our method, as evidenced by the semantic metrics and qualitative results. The results of Ours-M (EDM) and Ours-M (CD) are comparable to those of Luan$^\star$, which is also diffusion-based, but they incur substantially lower computational overhead, especially Ours-M (CD) that needs requires a single forward evaluation. The performance of Ours-M (CD) is slightly lower than that of Luan$^\star$. However, when considering semantic metrics, we observe that Ours-M (CD) only underperforms Luan$^\star$ in IoU for ground class , while significantly exceeding Luan$^\star$ in the IoU of all other classes. This indicates that the lag in geometric metrics is primarily due to the points of ground class. Regarding other classes which are more important for environmental perception, Ours-M (CD) achieves better results than Luan$^\star$, as also evidenced by the qualitative results. It is worth noting that the results of Radar-Diffusion$^\star$ which takes radar RCC as input are far inferior to those of Luan$^\star$ which uses RPC as input. We attribute this to the noisy and low-resolution nature of RCC, which makes it difficult to predict fine LiDAR-like point clouds when using them as control input. This observation further validates the effectiveness of our proposed multi-stage approach. We provide a more detailed analysis of this issue in the ablation study section.
    
    In summary, the quantitative results presented above demonstrate that our method outperforms traditional CFAR detector and other learning-based baselines in both 3D structure generation and semantic prediction.


    \subsection{Ablation Study}
    \label{sec:ablation}
    To evaluate the effectiveness of components and input modality in Sem-RaDiff, we conduct ablation experiments on the test set of the self-built dataset. All experiments are conducted under the same model configuration (Ours-M) and inference configuration (EDM), with the results shown in Table \ref{tab:ablation}.  

    \textbf{w/o SNR Enhancing}. We carry out this experiment by replacing multi-frame accumulated RCC obtained by parallel frame accumulation with single-frame RCC. The proposed parallel frame accumulation gains the performance for most metrics, especially for the IoU of wire class as well. This is attributed to the non-coherent accumulation which enhances the SNR of RCCs, with pronounced improvements for weak targets such as wires.  
    
    \textbf{w/o RCC Filtering}. In this experiment, we directly take RCC and RPC as the conditional input without Stage \uppercase\expandafter{\romannumeral1} prediction. As a result, the performance of our method Sem-RaDiff across multiple metrics, particularly with a significant drop in the IoU of wire class. This is due to the weak echo signals from wires in the RCC and the sparse wire points in the RPC. In contrast, wire target manifests as dense point in the coarse mask generated by Stage \uppercase\expandafter{\romannumeral1}, which provides rich structural and semantic information for Stage \uppercase\expandafter{\romannumeral2}, as shown in Fig.\ref{pic:architecture}. The results verify the effectiveness of Stage \uppercase\expandafter{\romannumeral1} within Sem-RaDiff.
    
    \textbf{Input Modality (RCC only and RPC only).} In our approach, both RCC and RPC are incorporated as network inputs. To investigate the individual contribution of each modality, separate experiments were conducted for RCC-only and RPC-only configurations. Under the former configuration, severe performance degradation occurs, with most metrics exhibiting the poorest results across all ablation settings. We posit this stems from the substantial modality gap between RCC and ground truth, making it challenging to predict fine-grained semantic point clouds solely from noisy and blurry  RCC. In contrast, the performance drop was considerably less pronounced when using only RPC, though there is still a notable performance loss compared to the baseline model, particularly in terms of Recall. This aligns with expectations since RPC provides only sparse point clouds as conditional input, leading to more missed detections, which RCC compensates for. However, we contend that the suboptimal performance of RCC-only case is partially attributable to dataset limitations. Due to flight authorization constraints, data collection is restricted to two agricultural fields, resulting in insufficient training sample diversity to effectively learn the joint distribution between RCC and ground truth. Future work will establish larger-scale datasets for further investigation.

    \subsection{System Efficiency}
    We evaluate the inference speed of our Sem-RaDiff on a Linux desktop with an Intel Core i7-10700 CPU and an NVIDIA GeForce RTX 4090 GPU with results listed in Table \ref{tab:model_efficiency}.  
    Note that FFT and CFAR are run on the radar sensor's processor, consuming no GPU resources, thus the whole pipeline includes parallel frame accumulation, Stage \uppercase\expandafter{\romannumeral1}, and Stage \uppercase\expandafter{\romannumeral1}. Both Stage \uppercase\expandafter{\romannumeral1} and Stage \uppercase\expandafter{\romannumeral2} are run with half-precision inference.
    As shown in Table \ref{tab:model_efficiency}, our proposed parallel frame accumulation scheme increases the computational speed of multi-frame radar cube accumulation by 40 times leveraging GPU, thereby promoting real-time onboard deployment. Ours-M and Ours-L achieve 13.6 FPS and 7.3 FPS on 4090 GPU, respectively. However, this performance still falls short of onboard deployment requirements due to limited compute resources in embedded systems (e.g., 4090 GPU offers 1321 AI TOPS\footnote{https://www.nvidia.com/en-us/geforce/graphics-cards/40-series/rtx-4090/}, while the Jetson Orin AGX provides 275 AI TOPS\footnote{https://www.nvidia.com/en-us/autonomous-machines/embedded-systems/jetson-orin/}).
	Nevertheless, Sem-RaDiff surpasses many 3D occupancy learning methods in inference speed\cite{sze2024real}, and they have demonstrated real-time performance on embedded AI chips after inference acceleration. This indicates the potential of Sem-RaDiff for onboard deployment after applying acceleration techniques such as model pruning and quantization, which we plan to explore in future work.

    \section{Conclusion}
    \label{sec:conclusion}
    In this paper, we propose Sem-RaDiff, a novel 3D mmWave radar semantic perception approach tailored for agricultural applications. To address the inherently noisy, blurry, and enormous nature of radar data, we introduce a coarse-to-fine framework that integrates discriminative and generative (diffusion) models, alongside parallel radar data accumulation and a sparse neural network. For validation, we build a dataset collected in real-world agricultural fields. Experimental results demonstrate that Sem-RaDiff outperforms existing methods in both 3D structure reconstruction and semantic prediction, while requiring significantly less computational overhead. It exhibits strong potential for real-time deployment on agricultural robots, serving as a viable alternative to optical sensors (e.g., cameras, LiDAR) in visually challenging conditions.
    
    In the future, we will explore onboard deployment of Sem-RaDiff and extend its application to downstream tasks such as autonomous navigation for agricultural robots. Additionally, we plan to establish and release a larger radar dataset of agricultural field scenes with more diverse scenarios and a greater variety of annotations to promote research on radar perception in agricultural settings.


    \bibliography{main_zrb}

\end{document}